\documentclass{article}

\usepackage[textwidth=6.25in, textheight=9in, centering]{geometry}

\usepackage{natbib}
\usepackage[utf8]{inputenc} %
\usepackage[T1]{fontenc}    %
\usepackage{hyperref}       %
\usepackage{url}            %
\usepackage{booktabs}       %
\usepackage{amsfonts}       %
\usepackage{amsmath}        %
\usepackage{amssymb}        %
\usepackage{microtype}      %
\usepackage{xcolor}         %
\usepackage{graphicx}       %
\usepackage{enumitem}       %
\usepackage{pifont}         %
\usepackage{placeins}       %

\newcommand{\R}{\mathbb{R}}
\newcommand{\E}{\mathbb{E}}

\newcommand{\diag}{\operatorname{diag}}
\newcommand{\erank}{\operatorname{erank}}

\newcommand{\rank}{\operatorname{rank}}

\newif\ifshowcomments
\showcommentsfalse

\graphicspath{{figures/}{../figures/}}

\title{Transforming Rank: How Architecture Navigates the Spectral Pathologies of Depth}

\author{%
  Katie Everett \\
  MIT CSAIL \\
  \texttt{everettk@csail.mit.edu}
}
\date{}  %

\begin{document}

\maketitle

\begin{abstract}
We investigate how each component of the Transformer feedforward block architecture
design determines how much rank survives across depth at initialization.
We reinterpret skip connections and normalization, long understood as controlling
magnitude, as mechanisms for preserving gradient rank across depth, since the very matrix
multiplications and nonlinear activations that make the network expressive also
reduce the rank.
We show that skip connections trade off rank collapse against ensemble-like behavior,
controlled by the relative scales of the branch and the skip: skip connections route
the gradient \emph{around} the residual branch, where rank is lost, rather than along
the long gradient paths that encourage the layers to compose.
The placement of the normalization layer controls this same tradeoff by setting the
branch-to-skip ratio across depth, unifying much of the normalization placement and
depth scaling literature, in particular why rank collapses for Post-Norm but plateaus
for Pre-Norm.
Other aspects of the architecture, like the two-matrix structure that expands and
contracts the width, use additional parameters to preserve the representation or
branch Jacobian rank.
The second matrix decorrelates a coherent mean spike that would grow across blocks with a
single matrix and uncentered activation, preventing the residual representation
from collapsing.
The width expansion between the two matrices keeps the branch Jacobian full rank:
applying the rank-reducing activation in this expanded space leaves enough directions
to span the original, at a width that follows a Marchenko--Pastur law.
The initialization rank of the input--output Jacobian predicts which networks train on CIFAR-10.
Taken together, we recast architecture design for deep networks as navigating an
intrinsic tradeoff among rank collapse, ensemble-like behavior, and parameter count.
Code is available at \url{https://github.com/everettk/transforming-rank-paper}.
\end{abstract}

\section{Introduction}
\label{sec:intro}

The modern Transformer feedforward block has a specific structure: each
block normalizes the residual stream, expands the hidden dimension by a factor of
roughly four through a pair of up- and down-projection matrices, applies an
elementwise activation between them, and adds the result back to the residual stream,
\[
  h_\ell = h_{\ell-1} + W_{\mathrm{down}}\,\sigma\!\left(W_{\mathrm{up}}\,\mathrm{Norm}(h_{\ell-1})\right).
\]

The overall structure of this feedforward block is almost exactly what the original Transformer paper
\citep{vaswani2017attention} proposed nearly a decade ago: what has changed in practice is the placement of
the normalization layer, from Post-Norm to Pre-Norm \citep{nguyen2019transformers, baevski2019adaptive}, and
the choice of activation, from ReLU through GELU \citep{hendrycks2016gelu} to gated units
\citep{shazeer2020glu}. Outside the feedforward block, there have been Transformer architecture innovations adopted widely
in practice, including mixture-of-expert models \citep{shazeer2017outrageously, fedus2022switch}
and modifications of attention \citep{dao2022flashattention, qiu2025gated}.
However, the design of the feedforward block itself has been exceptionally stable over time even as frontier models
have scaled up to hundreds of billions or more parameters.

Even at these scales, the depth of frontier models remains on the order of a hundred
layers \citep{llama3, deepseekv3}.
Training deep Transformers has challenges with training stability \citep{bachlechner2020rezero, zhang2019fixup},
diminishing returns from additional layers \citep{fahim2026depth, liu2026inversedepth, sun2025curseofdepth},
and signal propagation across depth \citep{noci2022signal}. One form of signal
degradation is rank collapse: the rank of the gradients and representations decreases
through the layers, until they span only a small number of directions in a
high-dimensional space \citep{feng2022rank}. In this work, we investigate how the architecture of the feedforward
block determines the effective rank of the gradients and representations in deep networks.

In a plain MLP, the input--output Jacobian, the gradient of the network output with respect to its
input, is a product of matrices: each layer contributes a weight matrix and a diagonal matrix of
activation derivatives. Since a product of matrices can only lose rank with each additional matrix
multiplication, the Jacobian rank decreases monotonically with depth. The literature on dynamical
isometry shows that the only way to avoid this rank loss in a plain MLP is for every per-layer
Jacobian to have all singular values near one \citep{saxe2014exact, pennington2017resurrecting}: this
requires that neither the weight matrices nor the activation derivatives reduce rank. In particular,
the activation function trades off expressivity against rank: an activation that preserves rank must
have a derivative near one in magnitude \citep{pennington2017resurrecting}; therefore, a nonlinearity
that makes the network expressive
also reduces the Jacobian rank \citep{murray2021activation}. Outside of dynamical isometry, where
typical neural networks operate, every layer of a plain MLP is a rank-reducing operation for the
Jacobian.

Skip connections \citep{he2016deep} and normalization layers \citep{ba2016layer, zhang2019root} are usually thought of as ways to keep the gradient and
representation magnitudes from vanishing or exploding with depth, whereas we reinterpret them as
mechanisms for \emph{bypassing} rank loss in the gradient. We show that skip connections preserve rank by
routing the gradient around the residual branch, where rank is lost, then restoring the rank through
the identity path around the branch that is trivially full-rank. The branch scale and initialization control how much of the
gradient flows through the residual branch versus around it, which comes with a tradeoff: at the
extreme, if all the gradient flowed around the branch, the network would act like an ensemble of
individual layers rather than a deep network whose layers compose. We further show that the placement
of the normalization layers controls this same ratio between the branch and skip, and therefore the tradeoff between rank collapse and
ensemble regimes; this perspective unifies many recent results on normalization placement and depth
scaling.

On the other hand, we show that the two-matrix structure in each feedforward block actually preserves
the rank of the residual branch Jacobian. The branch expands the hidden dimension with its first
matrix, applies the activation, and contracts back with the second matrix. Although the activation
still reduces rank, applying it in this expanded space rather than the model dimension lets enough
directions survive to keep the branch Jacobian full rank. In addition, the second matrix decorrelates
the residual branch means, so the representation does not collapse onto a single direction.

We argue there is a logical chain of architectural choices that navigate the rank tradeoff across depth, building up from a plain MLP to the modern
feedforward block. We measure the effective rank at initialization
of the input--output Jacobian, the residual stream representation, and the residual branch
Jacobian, and show that the rank of the input--output Jacobian predicts which networks train on CIFAR-10. Our
contributions are as follows:

\begin{itemize}[leftmargin=*, nosep, topsep=2pt]
  \item We show that skip connections preserve rank by routing the gradient around the
  rank-reducing branch through the full-rank identity path, creating a tradeoff between
  rank collapse and ensemble-like behavior, controlled by the relative magnitudes of the
  branch and skip paths (Section~\ref{sec:skip}).
  \item We show that the normalization placement determines the rank by controlling this
  same branch-to-skip ratio across depth: Post-Norm holds it constant and the rank collapses,
  while Pre-Norm lets it decay so the rank plateaus at a moderate level, at the expense of its
  deep layers shifting back toward the ensemble regime. The projection terms in the
  normalization operation contribute negligible rank loss compared to the branch-to-skip
  ratio. This same ratio controls a $\sqrt{L}$ depth factor in the initialization rank and
  unifies a range of results on normalization placement and depth scaling
  (Section~\ref{sec:norm-placement}).
  \item We show the two-matrix structure in the feedforward block preserves residual representation rank. With a
  single matrix and an uncentered activation, the rank of the residual representation
  collapses through a coherent mean spike whose cross-layer alignment equals the activation
  moment ratio $c_\ell = \mathbb{E}[\sigma]^2/\mathbb{E}[\sigma^2]$. The second matrix
  decorrelates this alignment to zero and prevents the collapse (Section~\ref{sec:coherence}).
  \item We show the width expansion between the two matrices preserves rank in the branch
  Jacobian. The activation function reduces the rank by masking or attenuating directions;
  applying this operation in the expanded space means enough directions survive to span the
  original space. We derive from a Marchenko--Pastur law the threshold $m/d = 1/p(\sigma)$ at
  which the branch Jacobian becomes full rank, where $p$ is the survival rate of the
  activation ($1/2$ for ReLU; $p = \mathbb{E}[\sigma'(z)^2]$ for smooth activations;
  Section~\ref{sec:width}).
\end{itemize}

In this work, we focus primarily on the initialization rank across depth in a network of
feedforward blocks. Related work studies the rank collapse that occurs in attention layers
\citep{dong2021attention, noci2022signal} and how rank evolves throughout training
\citep{martin2021implicit, galanti2025sgd}. Even at initialization, the rank behavior
across depth illustrates how the components of the feedforward block, though usually
understood in other terms, together navigate an intrinsic tradeoff among rank collapse,
ensemble-like behavior, and parameter count: skip connections and normalization route the
gradient around the rank-reducing branch, trading off rank collapse against ensemble-like
behavior, while the two-matrix structure and width expansion preserve the branch and
representation rank at the cost of additional parameters.

\begin{figure*}[!tb]
  \centering
  \includegraphics[width=0.334\textwidth]{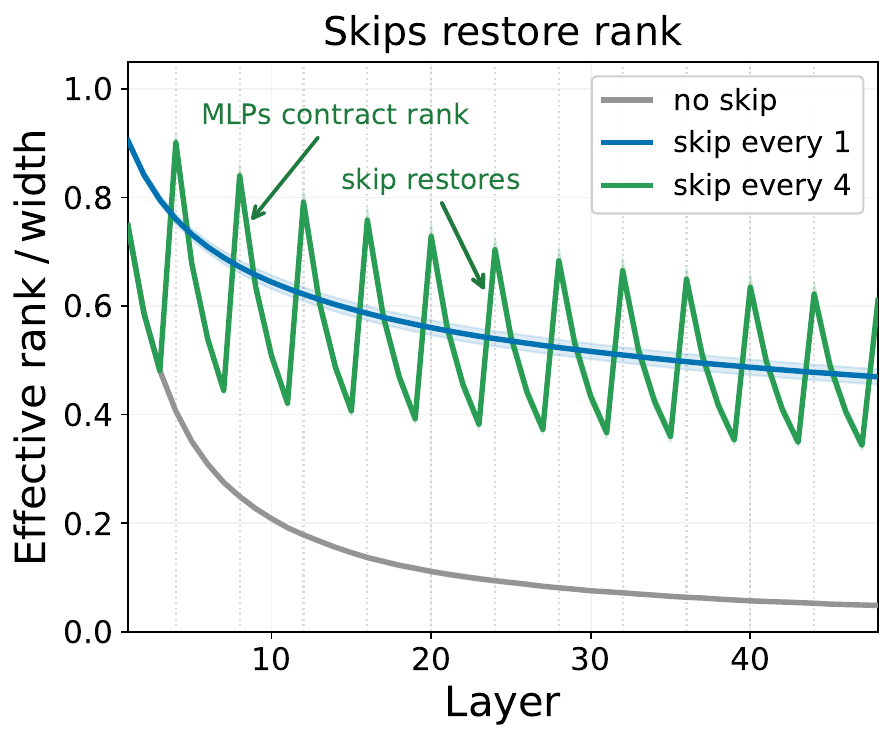}\hfill
  \includegraphics[width=0.316\textwidth]{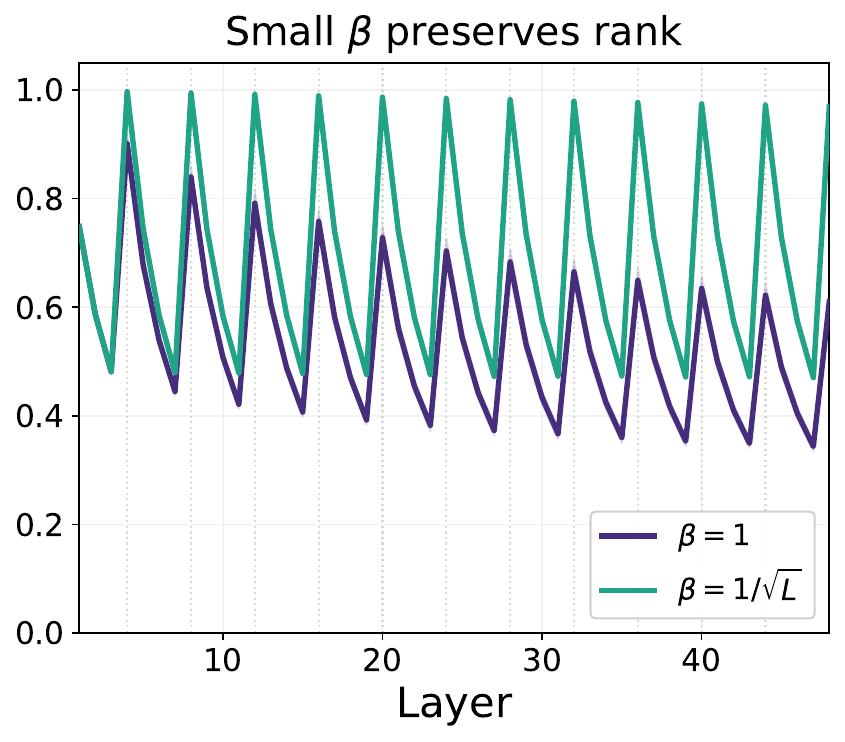}\hfill
  \includegraphics[width=0.316\textwidth]{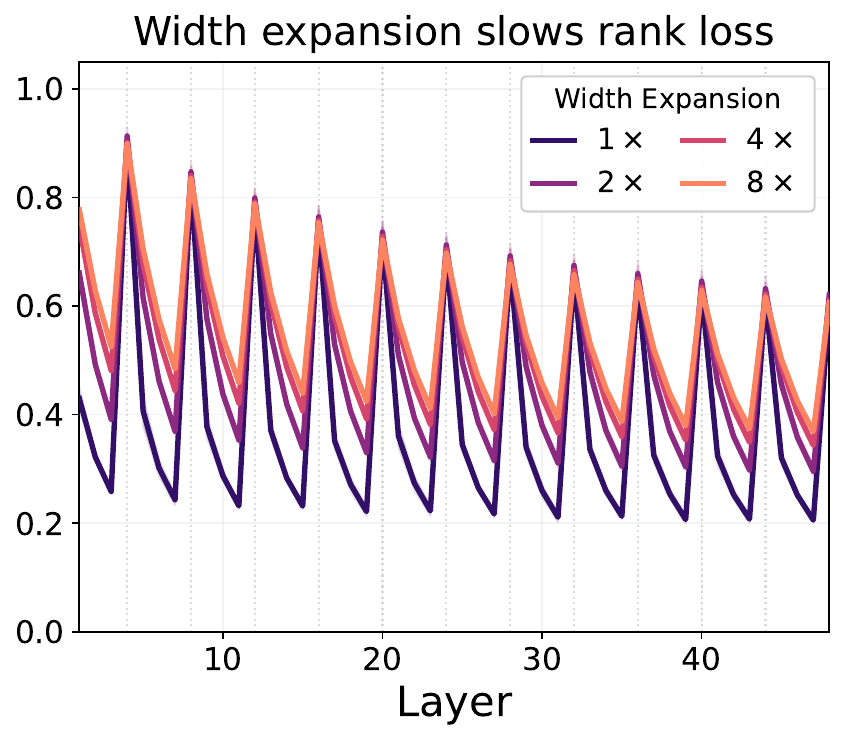}
  \caption{\textbf{Skip connections restore the effective rank that each
  feedforward block contracts; the residual scale $\beta$ and the width
  expansion control how much survives with depth.}
  Effective rank of the Jacobian $\partial h_\ell / \partial h_0$ up to each layer $\ell$,
  normalized by the width (model dimension) $d$, for a Pre-Norm ReLU residual
  MLP ($L = 48$, $d = 256$, $\beta = 1$, $\alpha = 1$, and $4\times$ width expansion unless
  otherwise stated; mean over 12 seeds).
  \emph{Left:} rank collapses without skips (gray); rank decays smoothly with a
  skip on every block (blue); spacing skips to every fourth block reveals the
  mechanism (green): rank contracts through the feedforward blocks and is
  restored at each skip. \emph{Center:} scaling the residual branch by
  $\beta = 1/\sqrt{L}$ reduces the magnitude of its contribution relative to the
  skip; since the skip (identity function) dominates, rank is preserved across
  depth, unlike at $\beta = 1$. \emph{Right:} residual branches with higher
  width expansion factors lose rank more slowly, raising the sawtooth troughs, a
  per-block conditioning effect analyzed in Section~\ref{sec:width}.}
  \label{fig:sawtooth}
\end{figure*}

\section{Setup}
\label{sec:setup}

We consider a residual network in which an input $h_0 \in \mathbb{R}^d$ is
transformed by $L$ residual blocks into $h_L$. Each residual block has the form
\begin{equation}
  h_\ell = h_{\ell-1} + \beta\, f(h_{\ell-1}), \qquad \ell = 1, \dots, L,
  \label{eq:residual-block}
\end{equation}
with branch scale $\beta$. The branch $f$ may itself include a normalization, an
elementwise activation $\sigma$, and one or two weight matrices; we write $M$ for
the number of branch matrices. By default the branch is the two-matrix
feedforward block, $f(h) = W_{\mathrm{down}}\,\sigma(W_{\mathrm{up}}\,\mathrm{Norm}(h))$,
with $W_{\mathrm{up}} \in \mathbb{R}^{m \times d}$ and
$W_{\mathrm{down}} \in \mathbb{R}^{d \times m}$, where $m/d$ is the expansion ratio.
Unless stated otherwise, the activation is ReLU, the normalization is RMSNorm in
the Pre-Norm placement,
$\beta{=}1$, and the weights are initialized as $\mathcal{N}(0, \alpha^2/\text{fan-in})$
with initialization scale $\alpha = 1$.

We track the rank, across depth, of three objects. The first is the
input--output Jacobian $\partial h_L/\partial h_0$, an ordered product over the
$L$ blocks,
\begin{equation}
  \frac{\partial h_L}{\partial h_0}
    = (I + \beta J_L)(I + \beta J_{L-1})\cdots(I + \beta J_1),
  \label{eq:io-jacobian}
\end{equation}
where $J_\ell := \partial f(h_{\ell-1})/\partial h_{\ell-1}$ is the branch Jacobian of
block $\ell$. The other two are the forward residual representation $Z_\ell$
(Section~\ref{sec:coherence}) and the branch Jacobian on its own, whose
normalization-free core is written $J_f$ (Section~\ref{sec:width}).

We report two spectral measures of a matrix $A$ with $N$ singular values
$\sigma_1 \ge \cdots \ge \sigma_N$. Our primary measure is the effective rank
\citep{roy2007effective}, the exponential of the spectral entropy,
\begin{equation}
  \erank(A) = \exp\!\Bigl(-\textstyle\sum_k p_k \ln p_k\Bigr),
  \qquad p_k = \sigma_k\big/\textstyle\sum_j \sigma_j,
  \label{eq:erank}
\end{equation}
which we report normalized by the model dimension $d$ as
$\erank(\partial h_L/\partial h_0)/d$; it equals $1$ when all singular values are
equal and drops toward $0$ as the spectrum collapses in a few directions. We
also use the stable rank
$\operatorname{st}(A) = \lVert A\rVert_F^2/\lVert A\rVert_2^2$, primarily in
Section~\ref{sec:coherence}. Both measures are scale-invariant and lie in $[1, N]$.
For Jacobian measurements taken at initialization, we sample a random Gaussian input
and evaluate each branch Jacobian at the resulting forward activations,
reporting means over independent samples of the weights and the input unless
stated otherwise.

\section{Skip Connections and Rank Restoration}
\label{sec:skip}

A plain MLP loses effective rank with depth due to repeated matrix
multiplications and nonlinearities: its per-layer Jacobians are each mildly
deficient in effective rank, and composing them compounds the loss across depth.
The only exception is networks satisfying the
dynamical isometry condition, which requires every
singular value of the input--output Jacobian to remain close to one, so that effective rank
is preserved at any depth \citep{saxe2014exact}. However, dynamical isometry in
neural networks is difficult to achieve in
practice, requiring orthogonal initialization and precise choices of activation
function and activation distribution maintained throughout training
\citep{pennington2017resurrecting, murray2021activation}.

Skip connections are often understood as an architectural solution to vanishing
gradient \emph{magnitude} across depth; we reinterpret them as a solution to
vanishing gradient \emph{rank} across depth. Gradients that flow through many MLP
layers inevitably lose rank, because the matrix multiplications and nonlinearities
that make the network expressive also reduce rank. Skip connections, in that sense,
are a workaround to this intrinsic rank loss: they preserve rank because they route
the gradient around the residual branch, where rank is lost, through an identity path
that is trivially full rank. This creates a \emph{sawtooth pattern} in the rank across depth if the
skips are spaced across multiple layers, where rank decreases through the MLPs and is
restored at each skip connection (Figure~\ref{fig:sawtooth}, left). In this section,
we argue that the deep residual architecture navigates a fundamental tradeoff: if all
of the gradient passed through the branch, the rank would collapse; if it all passed
through the skips, the network would act like an ensemble of individual layers rather
than a deep network whose layers compose.

\begin{figure*}[tb]
  \centering
  \includegraphics[width=\textwidth]{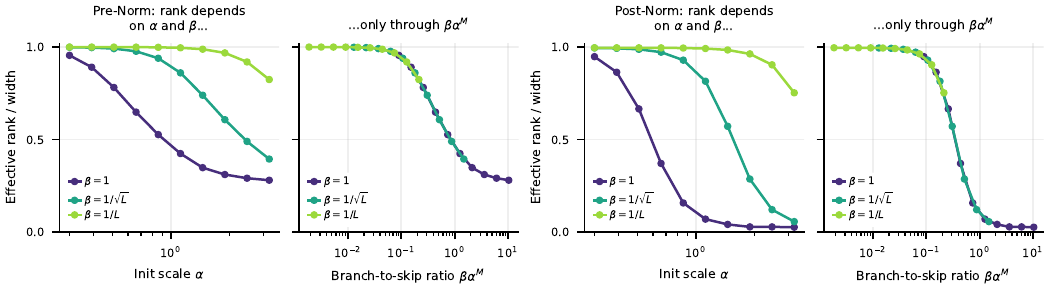}
  \caption{\textbf{The effective rank at initialization depends on the branch scale
  $\beta$ and the initialization scale $\alpha$ only through the nominal branch-to-skip
  ratio $\beta\alpha^{M}$ ($M$ the number of branch matrices), for both normalization
  placements.}
  Effective rank of the input--output Jacobian, normalized by the width, for a
  residual MLP at initialization ($d = 256$, $L = 48$, two square $d \times d$ branch
  matrices, single seed).
  \emph{(a)}~Pre-Norm: the rank depends on both the initialization scale $\alpha$
  and the branch scale $\beta$.
  \emph{(b)}~Plotted against $\beta\alpha^{M}$, the $\beta$-sweeps collapse onto a
  single curve.
  \emph{(c, d)}~The same holds for Post-Norm, but the rank collapses further at a
  given ratio.}
  \label{fig:rank-symmetry}
\end{figure*}

The standard residual block contains a scaled branch added to an identity skip
connection, $h_\ell = h_{\ell-1} + \beta f(h_{\ell-1})$. When the residual stream passes through each block untouched, as it does under
Pre-Norm (Appendix~\ref{app:skip}), the skip term must be the identity: scaling it by
$\lambda$ multiplies its contribution by $\lambda^{L-\ell}$, which explodes for
$\lambda > 1$ and vanishes for $\lambda < 1$ \citep{he2016identity}. The branch has two parameters that control its scale, the
branch scale $\beta$ and the initialization scale $\alpha$ of its weight matrices.
The Jacobian of a single block contains a skip term and a branch term; we define the
branch-to-skip ratio $r_\ell$ for each layer as the ratio of their Frobenius norms:
\begin{equation}
  \frac{\partial h_\ell}{\partial h_{\ell-1}}
  = \underbrace{I}_{\text{skip}} + \underbrace{\beta J_\ell}_{\text{branch}},
  \qquad
  r_\ell \equiv \frac{\|\beta J_\ell\|_F}{\|I\|_F}.
  \label{eq:branch-skip}
\end{equation}
At unit residual stream scale, $r_\ell = \sqrt{p}\,\beta\alpha^{M}$: the branch Jacobian
contributes one factor of $\alpha$ for each of the $M$ weight matrices, and a factor
$\sqrt{p}$ from the activation derivative, where $p = \mathbb{E}[\sigma'(z)^2]$
($p = 1/2$ for ReLU; for activations whose derivative is zero or $\pm 1$, this is the
survival rate in Section~\ref{sec:width}). We
will analyze the depth dependence of $r_\ell$ in Section~\ref{sec:norm-placement}. For residual networks with
positively homogeneous activations, there are equivalence classes over the initialization
scale $\alpha$ and the branch scale $\beta$ in which networks with the same value of
$\beta\alpha^{M}$ compute the same function at initialization. This equivalence persists
throughout training if the optimizer hyperparameters are rescaled accordingly for each
matrix (Appendix~\ref{app:skip-symmetry}).
In particular, we show that the effective rank at initialization depends on $\beta$ and
$\alpha$ only through $\beta\alpha^{M}$, for both Pre- and Post-Norm
(Figure~\ref{fig:rank-symmetry}). A larger $\beta\alpha^{M}$ sends more of the gradient
through the branch rather than the skip, so the rank collapses faster with depth.

\begin{figure*}[tb]
  \centering
  \includegraphics[width=\textwidth]{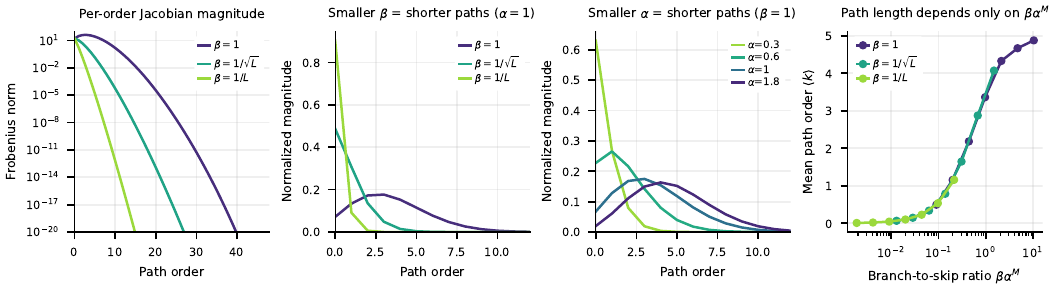}
  \caption{\textbf{Short paths (mostly skip connections) dominate the input--output
  Jacobian magnitude; a smaller branch scale $\beta$ or initialization scale $\alpha$
  shifts it toward even shorter paths. The path lengths depend only on the
  nominal branch-to-skip ratio $\beta\alpha^{M}$.}
  Path decomposition of a Pre-Norm residual MLP at initialization ($d = 256$,
  $L = 48$, $4\times$ width expansion; mean over 8 seeds).
  \emph{(a)}~Per-order Jacobian magnitude: the Frobenius norm of the order-$k$ paths,
  those through exactly $k$ feedforward blocks and $L-k$ skip connections; order $0$
  only travels through the skip.
  \emph{(b)}~The same magnitudes normalized to sum to one; a smaller branch scale
  $\beta$ concentrates the magnitude on shorter paths.
  \emph{(c)}~A smaller initialization scale $\alpha$ also shifts the magnitude toward
  shorter paths.
  \emph{(d)}~Mean path order $\langle k \rangle$, sweeping the initialization scale
  $\alpha$ at each of three branch scales $\beta$; $\langle k \rangle$ depends only on
  the branch-to-skip ratio $\beta\alpha^{M}$ ($M$ the number of branch matrices), so
  all points fall on a single curve.}
  \label{fig:paths}
\end{figure*}

Prior work shows that deep residual networks behave like ensembles of shallow
networks \citep{veit2016residual, liu2026inversedepth}. \citet{veit2016residual}
decompose a residual network forward pass into $2^L$ paths that take either the skip
or the branch at each block. We apply a similar expansion to the input--output
Jacobian, where the skip contributes the identity and the branch contributes
$\beta J_\ell$ to the product. Grouping the paths by their order $k$, the number of
blocks at which a path takes the branch rather than the skip, gives
\begin{equation}
  \frac{\partial h_L}{\partial h_0} = \sum_{k=0}^{L} \beta^k S_k, \qquad
  S_k = \sum_{1 \le \ell_1 < \cdots < \ell_k \le L} J_{\ell_k} \cdots J_{\ell_1},
  \label{eq:path-decomp}
\end{equation}
where each order-$k$ path carries weight $(\beta\alpha^{M})^k$. We say a path is
\emph{short} if it passes through few feedforward blocks (small $k$) and runs mostly
through the skip connections; a \emph{long} path (large $k$) passes through many
feedforward blocks.
Because the path weight is raised to the order $k$, small changes in $\beta$ and
$\alpha$ are amplified in the long paths. Small changes in the ratio therefore
induce significant shifts in the distribution over path orders
(Figure~\ref{fig:paths}): at a small ratio the short paths dominate, which encourages
the network to behave like an ensemble of shallow subnetworks more than a deep
network whose layers compose. This same ratio also controls whether the rank shows
a flat or a decaying sawtooth pattern: at $\beta = 1/\sqrt{L}$ each skip nearly
restores the rank lost in the preceding block, whereas at $\beta = 1$ it restores
only part and rank declines with depth (Figure~\ref{fig:sawtooth}, center).

Our decomposition lets us reinterpret the downscaling of the residual branch,
often recommended in the depth-scaling literature, as a way to bypass rank
collapse but possibly at the expense of composition. Multiple theoretical
perspectives all propose $\beta=1/\sqrt{L}$ scaling, motivated by gradient
stability and an expressive infinite-depth kernel \citep{hayou2021stable},
feature learning and hyperparameter transfer \citep{yang2023depth,
bordelon2023depthwise}, and preventing rank collapse across the sequence
dimension \citep{noci2022signal}. \citet{tarnowski2019dynamical} show that
residual networks with any activation function achieve dynamical isometry when
the branch is downscaled sufficiently. In practice, though, many frontier models
keep $\beta=1$ \citep{llama3, deepseekv3, gemma2, qwen3, gptoss2025, kimik2} or
similarly rescale the skip
connection \citep{minimax2025}, suggesting the stronger branch contribution may bring
empirical benefits on downstream tasks that downscaling would sacrifice.

A network that does not downscale the branch needs normalization layers to
control the growing magnitude of the residual stream; as the next section will
show, the normalization placement also determines how the branch-to-skip ratio,
and therefore the rank, evolves with depth. At $\beta=1$ the stream grows because
the branch contributions accumulate rather than cancel at initialization
(Appendix~\ref{app:skip}). Some methods avoid normalization entirely by
initializing the branch near zero \citep{bachlechner2020rezero, de2020skipinit,
zhang2019fixup}: a small ratio limits the magnitude growth that normalization
would otherwise control, at the same possible cost in composition as downscaling
the residual branch with small $\beta$.

\section{Normalization Placement and the Branch-to-Skip Ratio}
\label{sec:norm-placement}

A normalization layer can be placed at three locations in a residual block: the branch
input (Pre-Norm), the branch output (which we call Output-Norm), or the stream after the
residual addition (Post-Norm) \citep{kim2025peri}. The original Transformer used Post-Norm \citep{vaswani2017attention},
\begin{equation}
  h_\ell = \mathrm{Norm}\bigl(h_{\ell-1} + \beta\, f(h_{\ell-1})\bigr),
  \label{eq:postnorm}
\end{equation}
but the canonical architecture today uses Pre-Norm \citep{llama3,deepseekv3}:
\begin{equation}
  h_\ell = h_{\ell-1} + \beta\, f(\mathrm{Norm}(h_{\ell-1})).
  \label{eq:prenorm}
\end{equation}
The optimal placement of normalization layers is still debated, with recent work
proposing that Post-Norm can be viable in deep networks
\citep{wang2024deepnet,liu2024branchnorm,chen2026post} or that normalization is
beneficial at other locations, including several per block or different placements at
different depths \citep{ding2021cogview,shleifer2021normformer,wang2023magneto,kim2025peri,li2025mixln,gemma2,gemma3}.

\begin{figure}[t]
  \centering
  \includegraphics[width=\textwidth]{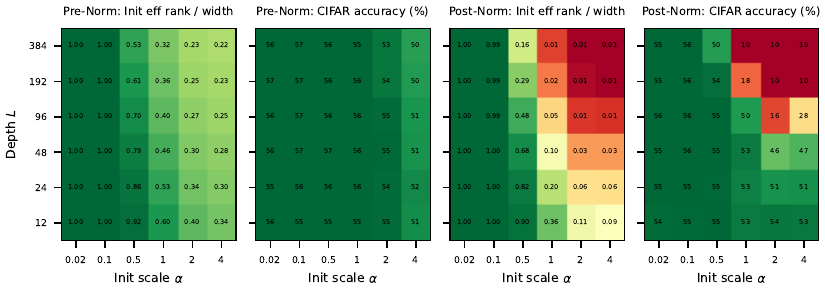}
  \caption{\textbf{The initialization effective rank predicts CIFAR-10 trainability.}
  Heatmaps over initialization scale $\alpha$ and depth $L$, for residual networks with a two-matrix
  ReLU branch (both matrices $d\times d$, $d{=}256$) and $\beta{=}1$.
  \textbf{(a,c)}~Effective rank of the input--output Jacobian at initialization.
  \textbf{(b,d)}~Best CIFAR-10 test accuracy over a learning-rate sweep with Adam.
  For Pre-Norm \textbf{(a,b)}, the rank stays above $0.2$ and training succeeds across the
  entire grid. For Post-Norm \textbf{(c,d)}, the rank collapses at large $\alpha$ and
  depth, and accuracy falls to chance ($10\%$) in the same region.}
  \label{fig:norm-trainability}
\end{figure}

We consider normalization operations at each location that subtract the mean, rescale
the root-mean-square of the input, or do both.\footnote{This setup includes RMSNorm
(which only rescales) and LayerNorm (which does both), but not BatchNorm, which
normalizes across the batch rather than within a vector \citep{ioffe2015batch}.}
This section shows that the normalization placement and operation determine the effective
rank through how the branch-to-skip ratio (Eq.~\ref{eq:branch-skip}) changes across depth: a
post-residual rescaling holds the ratio constant and the rank collapses,
whereas every other choice lets the ratio decay and the rank plateaus, provided a
normalization fixes the scale of the branch. The rank measurements in this section use a
single sample of the weights and the input, and the operations we compare share the same
sample; each cell of the CIFAR heatmaps (Figure~\ref{fig:norm-trainability}) trains one
network per learning rate.

We first rule out a potential explanation for why the normalization placement affects the
rank: prior work observes the normalization Jacobian contains a projection term that
removes directions from the gradient \citep{xu2019understanding,emadi2026exact}. Deep
Post-Norm networks show collapsed input--output Jacobian spectra already at initialization
\citep{bachlechner2020rezero}. The mean
subtraction operation removes the all-ones direction $\mathbf{1}$, and the rescaling
operation removes the current stream direction
$\hat{x} = h/\|h\|$ (with the normalization gain at its standard initialization of one):
\begin{equation}
  \frac{\partial}{\partial h}\bigl(h - \mu(h)\mathbf{1}\bigr)
    = \underbrace{I - \tfrac{1}{d}\mathbf{1}\mathbf{1}^{\top}}_{\text{removes } \mathbf{1}},
  \qquad\quad
  \frac{\partial}{\partial h}\,\frac{h}{\mathrm{RMS}(h)}
    = \frac{\sqrt{d}}{\|h\|}\,
      \underbrace{\bigl(I - \hat{x}\hat{x}^{\top}\bigr)}_{\text{removes } \hat{x}}.
  \label{eq:norm-proj}
\end{equation}
The placement changes where the
projection occurs relative to the identity. In the Post-Norm block Jacobian,
$J_{\mathrm{Norm}}(I + \beta J_f)$, every gradient path passes \emph{through} the
projection; in the Pre-Norm block Jacobian, $I + \beta J_f J_{\mathrm{Norm}}$, the
identity carries the gradient \emph{around} it \citep{emadi2026exact}. If the projection
removes directions from the gradient, and the identity bypass protects the rank in
Pre-Norm, then the projection could explain why the rank collapses in Post-Norm, where no
bypass exists.

\begin{figure}[t]
  \centering
  \includegraphics[width=\textwidth]{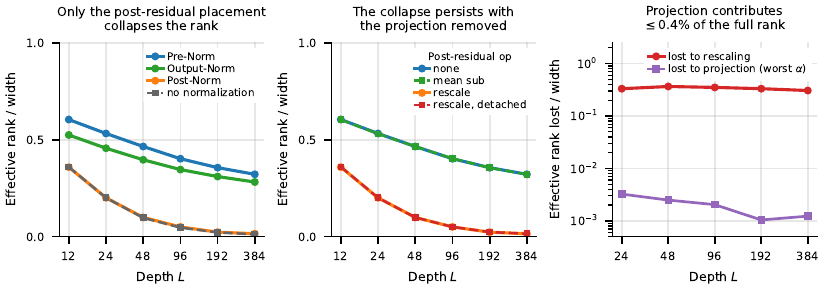}
  \caption{\textbf{The projection terms in the normalization Jacobian are not the cause of
  the rank collapse.} All panels: residual networks at initialization
  with a two-matrix ReLU branch (both matrices $d\times d$, $d{=}256$), $\beta{=}1$; effective
  rank of the input--output Jacobian, normalized by width. Panels (a) and (b) fix $\alpha{=}1$.
  \textbf{(a)}~Only the post-residual placement collapses rank. A normalization at either branch
  placement, the input (Pre-Norm) or the output (Output-Norm), keeps a floor; the post-residual
  placement (Post-Norm) collapses at the same rate as no normalization at all.
  \textbf{(b)}~The collapse persists with the projection removed. A detached rescaling
  (fixes the scale, removes no direction) collapses identically
  to RMSNorm; mean subtraction (removes a direction, does not rescale) keeps a floor. Every
  configuration uses a Pre-RMSNorm; we vary only the post-residual operation, and the rescaling
  collapse does not depend on it (Figure~\ref{fig:norm-nopre-control}).
  \textbf{(c)}~The projection's contribution to rank loss is negligible. The plot splits the
  rank a post-residual RMSNorm removes into a rescaling part and a projection part
  (Appendix~\ref{app:norm-control}): the rescaling accounts for essentially all of it, the
  projection at most $0.4\%$ of the full rank, across $\alpha\in\{0.5,1,2\}$.}
  \label{fig:norm-mechanism-dial}
\end{figure}

Our experiments present two pieces of evidence that rule out the projection as the
cause of the rank collapse.
Our first experiment is the placement comparison in Figure~\ref{fig:norm-mechanism-dial}a: a
normalization at the branch input or output keeps a rank floor, while Post-Norm loses rank
at the same rate as a network with no normalization at all. A network with no normalization removes no directions, which indicates that the
projection is not necessary to induce the collapse. Our second experiment is an ablation that removes the projection from the Jacobian
while changing nothing else. The
projection term in the RMSNorm Jacobian comes from differentiating through the stream
scale. We stop the gradient through that scale to obtain a \emph{detached rescaling}: the
forward pass is unchanged, while the Jacobian becomes a multiple of the identity that
removes no direction. This detached rescaling collapses the rank just like ordinary RMSNorm
(Figure~\ref{fig:norm-mechanism-dial}b). When we split the rank that a post-residual
RMSNorm removes into a rescaling part and a projection part, the projection accounts for at
most $0.4\%$ of the full rank, across initialization scales
(Figure~\ref{fig:norm-mechanism-dial}c; Appendix~\ref{app:norm-control}). From this, we
conclude that the post-residual normalization does not itself \emph{remove} rank from
the gradient; the placement simply fails to \emph{preserve} it the way that Pre-Norm does.

\begin{figure}[t]
  \centering
  \includegraphics[width=\textwidth]{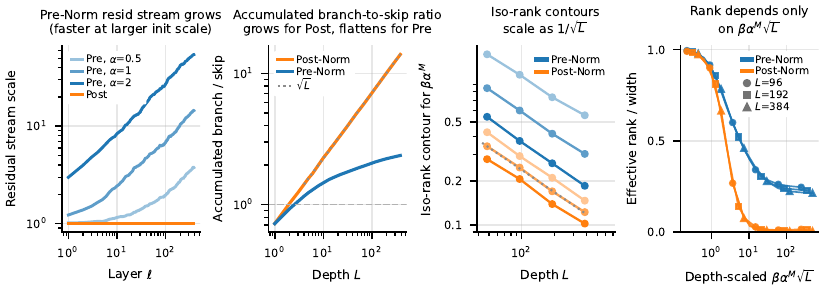}
  \caption{\textbf{The effective rank depends only on the product of the branch-to-skip ratio
  $\beta\alpha^{M}$ (exact symmetry) and a $\sqrt{L}$ depth factor (empirical).} Post-Norm
  collapses to zero, while Pre-Norm plateaus at a moderate level. All panels at initialization
  with a two-matrix ReLU branch (both matrices $d\times d$, $d{=}256$), $\beta{=}1$; $\beta\alpha^{M}$ is the nominal branch-to-skip
  ratio ($M$ branch matrices).
  \textbf{(a)}~The norm placement controls the branch-to-skip ratio at each depth. The Pre-Norm
  residual stream grows with depth, faster at larger initialization scale $\alpha$, so the ratio
  decays; Post-Norm rescales the residual stream to unit scale, which holds the ratio constant.
  \textbf{(b)}~The accumulated branch norm relative to the skip grows as $\sqrt{L}$ for Post-Norm
  (a random walk of constant-magnitude increments) and flattens for Pre-Norm (a random walk of
  shrinking increments). At $L{=}384$ the accumulated branch norm reaches $14\times$ the skip for Post-Norm
  and $2.4\times$ for Pre-Norm.
  \textbf{(c)}~The rank is an approximate function of $\beta\alpha^{M}\sqrt{L}$: for both
  placements, the iso-rank contours (rank levels $0.4$, $0.5$, $0.6$, lighter to darker) scale as
  $1/\sqrt{L}$.
  \textbf{(d)}~Plotted against $\beta\alpha^{M}\sqrt{L}$, the curves for all depths fall onto
  one curve per placement, confirming the rank depends on this combination alone.}
  \label{fig:norm-law}
\end{figure}

\textbf{Instead, we show that the normalization placement controls how the branch-to-skip
ratio changes across depth, which in turn determines the effective rank.} For Pre-Norm and
Post-Norm, with the branch scale $\beta$ and initialization scale $\alpha$ fixed across
layers, the branch-to-skip ratio varies with depth only through the scale of the residual
stream. Throughout this section we track its nominal form $\tilde r_\ell \equiv r_\ell / \sqrt{p}$,
dropping the constant activation-derivative factor that is shared by every block and both
placements (Section~\ref{sec:skip}). In Pre-Norm, the branch acts on a normalized input,
so its Jacobian contains the factor $\sqrt{d}/\|h_\ell\|$ from the normalization
(Eq.~\ref{eq:norm-proj}), giving $\tilde r_\ell = \beta\alpha^{M}\sqrt{d}/\|h_\ell\|$: the
nominal ratio $\beta\alpha^{M}$ divided by the stream scale $\mathrm{rms}(h_\ell)$. The Pre-Norm
stream grows with depth, since each block adds a branch output of roughly fixed magnitude
that is nearly orthogonal to the stream. The scale grows as $\|h_\ell\| \propto \sqrt{\ell}$
(Figure~\ref{fig:norm-law}a; Appendix~\ref{app:norm-law}) and the ratio decays as
$1/\sqrt{\ell}$. In Post-Norm, the rescaling resets the stream to unit scale at every
block, and the ratio stays fixed at $\beta\alpha^{M}$. With no normalization at all, the
ratio is also constant, but for a different reason: a degree-one homogeneous branch Jacobian does not
depend on the scale of its input, so the growing stream leaves the ratio unchanged.

To see why the branch-to-skip ratio determines the rank, recall from
Section~\ref{sec:skip} that the operations that reduce rank are in the residual branch,
and that the skip connection routes the gradient around them. Since the ratio controls how
much of the gradient passes through these rank-reducing operations compared to the skip,
the rank lost at each layer depends on this ratio. When the ratio decreases with depth, later blocks lose less and
less rank, and the rank levels off. When the ratio is constant, every block keeps losing
rank, and the rank collapses. This resolves the coincidence in Figure~\ref{fig:norm-mechanism-dial}a:
Post-Norm and no normalization lose rank at the same rate because both hold the
branch-to-skip ratio constant across depth.

We now measure how the effective rank depends on the branch-to-skip ratio and the depth
together. Recall from Section~\ref{sec:skip} that networks with the same $\beta\alpha^{M}$
compute the same function at initialization, and that the effective rank depends on
$\beta$ and $\alpha$ only through this quantity, for both Pre-Norm and Post-Norm
(Figure~\ref{fig:rank-symmetry}). To measure the depth dependence, we extract iso-rank
contours: at each depth $L$, the value of $\beta\alpha^{M}$ that holds the effective rank
at a fixed level. \textbf{We find that the effective rank at initialization depends
only on the product of the branch-to-skip ratio $\beta\alpha^{M}$ and a $\sqrt{L}$ depth
factor: for both Pre-Norm and Post-Norm, the iso-rank contours scale as $1/\sqrt{L}$
(Figure~\ref{fig:norm-law}c; measured slope $-0.50$).} Plotted against
$\beta\alpha^{M}\sqrt{L}$, the rank at every depth falls onto one curve for Pre-Norm and
another for Post-Norm (Figure~\ref{fig:norm-law}d): the Post-Norm curve collapses to zero,
whereas the Pre-Norm curve approximately plateaus at a moderate level.

We next give a heuristic scaling argument for the $\sqrt{L}$ depth factor, using the path
decomposition in Section~\ref{sec:skip} (Eq.~\ref{eq:path-decomp}). There are $\binom{L}{k}$
paths of order $k$, one for each choice of the $k$ blocks where the path passes through the
branch instead of the skip. For Post-Norm, each branch visit contributes a factor of
$\beta\alpha^{M}$ to a path's norm relative to the skip (we drop the post-residual norm
Jacobians, whose projections empirically contribute at most $0.4\%$ of the full rank). At
initialization, the branch Jacobians of different blocks are nearly independent, so the
paths of each order add like a random walk, to norm
$\sqrt{\binom{L}{k}}\,(\beta\alpha^{M})^{k} \approx (\beta\alpha^{M}\sqrt{L})^{k}/\sqrt{k!}$.
Each additional branch visit multiplies the number of available paths by roughly $L$, so one
factor of $\sqrt{L}$ comes with each factor of $\beta\alpha^{M}$. Every order of the
expansion then depends on $\alpha$, $\beta$, and $L$ only through $\beta\alpha^{M}\sqrt{L}$.
The effective rank is scale-invariant, so it is a function of this single variable, and the
iso-rank contours scale as $1/\sqrt{L}$.

The order-one paths form the accumulated branch in Figure~\ref{fig:norm-law}b, where $L$
randomly oriented terms of magnitude $\beta\alpha^{M}$ add like a random walk to
$\beta\alpha^{M}\sqrt{L}$ for Post-Norm. For Pre-Norm the growing stream shrinks the
per-layer ratios as $1/\sqrt{\ell}$, and each order of the expansion is again a function of
$\beta\alpha^{M}\sqrt{L}$; the accumulated branch flattens instead of growing
(Figure~\ref{fig:norm-law}b). The rank begins to decrease where the branch orders reach the
scale of the skip, around $\beta\alpha^{M}\sqrt{L} \sim 1$ (the order-one line in
Figure~\ref{fig:norm-law}b). For a more detailed argument, see Appendix~\ref{app:norm-law},
where we state the assumptions behind each step and validate them empirically, including on
GPT-2 Small and Medium both at initialization and after training.

This argument tells us that the rank depends on $\beta\alpha^{M}\sqrt{L}$, but not how; we
measure the shape of each curve empirically (Figure~\ref{fig:norm-law}d). For Post-Norm, the rank collapses to zero, since the accumulated
branch grows as $\sqrt{L}$ and dominates the skip. For Pre-Norm, the accumulated branch stays
comparable to the skip, so the rank plateaus at a moderate level.

The initialization rank predicts whether these networks train. We train on CIFAR-10 across
a grid of initialization scale $\alpha$ and depth $L$ for both placements at $\beta{=}1$, taking the
best test accuracy over an Adam learning-rate sweep (Figure~\ref{fig:norm-trainability}).
We note that CIFAR-10 is an easy task whose accuracy saturates even with a shallow network,
so this serves as a coarse test of trainability: it tests whether a configuration trains at
all, rather than how it benefits from depth. We see that the networks fail to train in the
configurations where the initialization rank collapses: when the rank collapses all the way
to 1--2\% effective rank over width, the CIFAR accuracy drops toward chance, whereas more
moderate rank collapse harms accuracy by several percentage points. Moreover, as depth
increases, the learning-rate basin narrows: the maximum stable learning rate decreases with
depth as rank collapse begins (Appendix~\ref{app:norm-lr}), whereas the CIFAR accuracy failures
occur at larger depths once the rank has fully collapsed. In this sense, the narrowing
learning-rate basin acts as an early signal of a trainability problem, before any harm to
the final accuracy. We include results for Output-Norm in Appendix~\ref{app:norm-slots}, which are
similar to Pre-Norm across the grid.

Our results unify a variety of methods in the normalization placement literature by
asking how each changes the branch-to-skip ratio across depth. The Jacobian rank collapse
is specific to the post-residual placement. A branch-input or branch-output placement
avoids normalizing the residual stream, so the skip remains an identity, the stream grows
with depth, and the branch-to-skip ratio decays. Architectures whose per-block
normalizations stay off the residual stream
\citep{ding2021cogview,shleifer2021normformer,wang2023magneto,kim2025peri} avoid the
problem. One line of work aims to revive Post-Norm at depth by up-scaling the skip inside
the normalization and/or down-scaling the branch initialization
\citep{wang2024deepnet,chen2026post}; this shrinks the branch relative to the skip, which
in our notation reduces $\beta\alpha^{M}$ and means a higher initialization rank at every
depth (Figure~\ref{fig:norm-law}d). Additional work reduces how many post-residual
normalizations the gradient must pass through \citep{takase2023b2t,wang2026spannorm}.
Finally, the $1/\sqrt{L}$ branch downscaling in Section~\ref{sec:skip} coincides with
our measured iso-rank contours: $\beta{=}1/\sqrt{L}$ holds
$\beta\alpha^{M}\sqrt{L}$ constant, so it is precisely the scaling at which the
initialization rank is depth-invariant, which may be part of why it works well.

This section has shown that the normalization placement and operation determine the Jacobian
effective rank at initialization through a single quantity, the branch-to-skip ratio across
depth; the projection term in the normalization Jacobian contributes negligible rank loss.
The placement determines whether the normalization rescales the residual stream: at the
branch input or output, the skip stays an identity, the stream grows, and the ratio decays
with depth. Among configurations where a normalization fixes the scale of the branch, the rank
collapses only under a post-residual rescaling, the one combination
where the branch-to-skip ratio stays constant across depth; every other placement and
operation lets the ratio decay, and the rank keeps a floor. Quantitatively, the rank depends
on the ratio and the depth only through $\beta\alpha^{M}\sqrt{L}$, and the initialization
rank predicts which configurations train.

Recall from Section~\ref{sec:skip} that explicitly downscaling the branch with small values
of $\alpha$ or $\beta$ preserves the rank at a possible cost in composition: short paths
through primarily skip connections become dominant, and the network behaves more like an
ensemble of shallow subnetworks. Pre-Norm applies this same downscaling implicitly: the
stream grows while each branch output keeps a fixed magnitude, so the ratio decays as
$1/\sqrt{\ell}$ and the deep blocks become small perturbations of the identity. Post-Norm
instead holds the ratio constant, so its blocks keep transforming the stream and keep losing
rank. \textbf{In Pre-Norm, the rank survives for the same reason the deep layers become
ineffective: a near-identity block neither removes directions from the gradient nor adds
computation beyond contributing its individual layer to the ensemble.} In trained Pre-Norm
language models, the deep layers indeed contribute little: they can be pruned individually
with minimal performance loss, and their block Jacobians approach the identity
\citep{sun2025curseofdepth}. In other words, the decaying branch-to-skip ratio suggests a
mechanism for this ineffectiveness, but it also explains the benefit: the rank
stays moderate, and the network still trains at depths where Post-Norm fails
(Figure~\ref{fig:norm-trainability}).

\FloatBarrier

\begin{figure*}[t]
  \centering
  \includegraphics[width=0.242\textwidth]{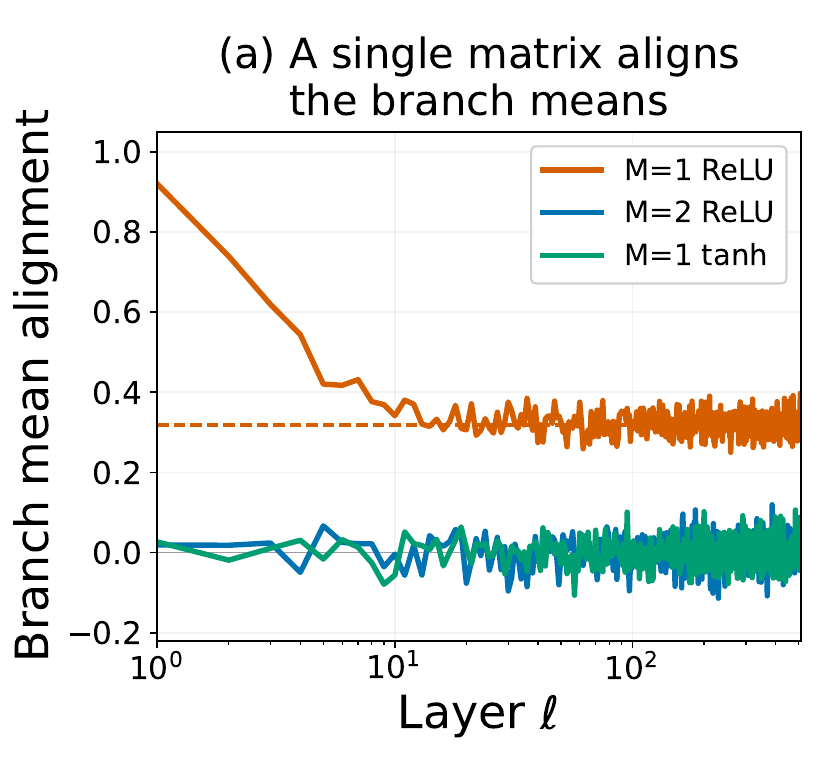}\hfill
  \includegraphics[width=0.242\textwidth]{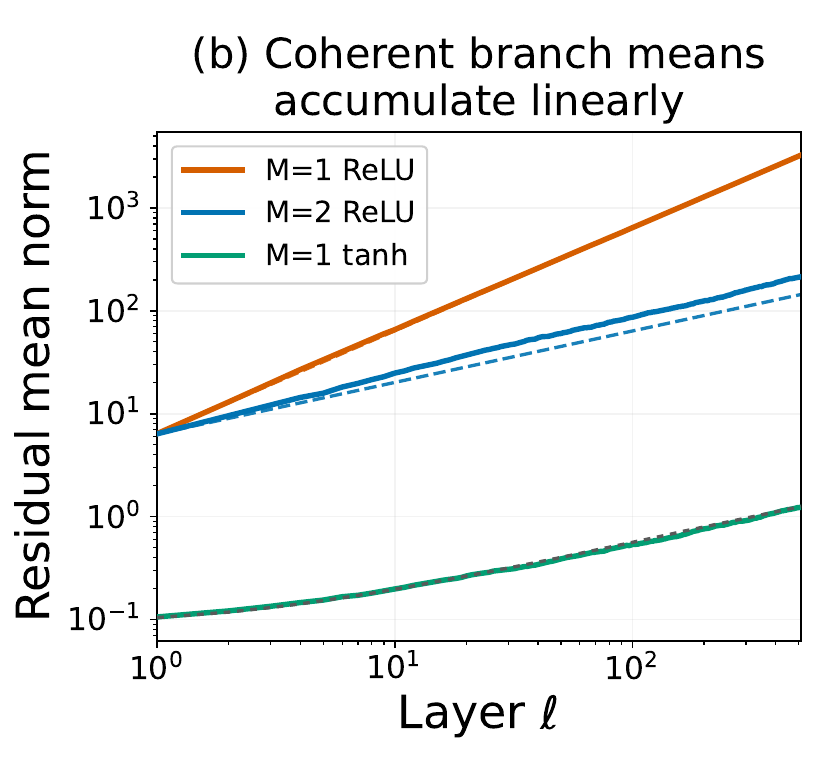}\hfill
  \includegraphics[width=0.242\textwidth]{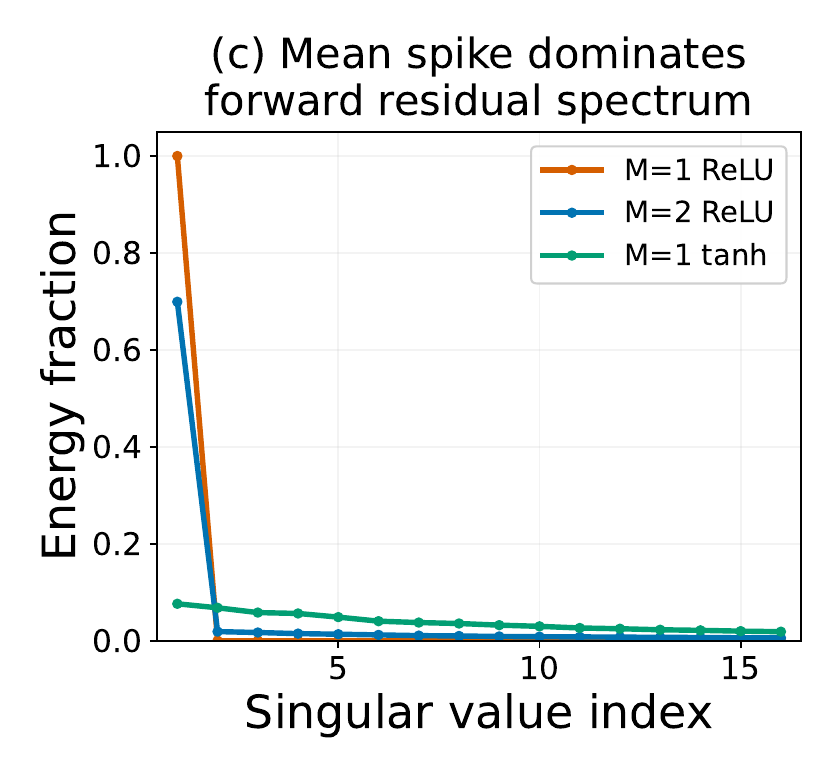}\hfill
  \includegraphics[width=0.242\textwidth]{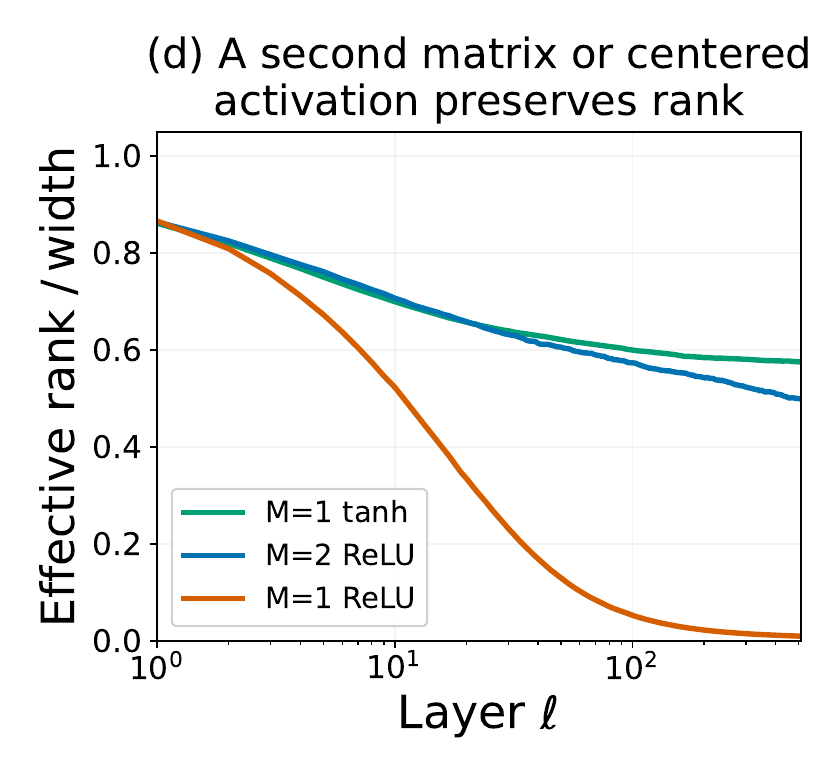}
  \caption{\textbf{A single uncentered branch matrix grows a mean spike that
  collapses the forward residual stream rank; a second matrix or a centered
  activation prevents the collapse.} We track the residual stream representation
  $Z_\ell$, the matrix whose rows are the representations of $n$ probe inputs at
  layer $\ell$, in a depth-$L$ network at initialization (Pre-RMSNorm,
  $\beta{=}1$, $d{=}256$, $L{=}512$, 3 seeds).
  (a) The mean of $Z_\ell$ over inputs is the sum $m_1 + \cdots + m_\ell$ of the
  per-layer branch means; we plot the alignment $\cos(m_\ell, m_{\ell+1})$ of
  consecutive terms. A single
  uncentered matrix aligns them at the coherence constant $c_\ell$ ($1/\pi$ for
  ReLU, dashed); a second matrix decorrelates them; tanh's branch means are
  zero, so there is nothing to align.
  (b) The norm of that sum grows linearly when the terms align ($\propto\ell$
  guide) and as a random walk when they do not ($\propto\sqrt{\ell}$). tanh's
  true mean is zero; its curve sits on the gray $n$-sample noise floor
  (Appendix~\ref{app:coherence}).
  (c) The spectrum of $Z_L$ at the final layer (top $16$ of $d{=}256$ singular
  values): the top singular value carries the mean, with energy fraction $1.00$
  for M=1 ReLU (a rank-1 spike) and $0.70$ for M=2 ReLU (a full tail remains);
  tanh has no spike, and its top direction holds only $0.08$, barely above the
  rest of its spectrum.
  (d) The effective rank of $Z_\ell$ collapses for M=1 ReLU and stays a finite
  fraction of the width for M=2 ReLU and tanh.}
  \label{fig:coherence}
\end{figure*}

\section{The Two-Matrix Branch and the Coherent Mean Spike}
\label{sec:coherence}

Transformer feedforward blocks place two weight matrices inside every residual
branch. The previous sections might suggest the opposite design: if skip
connections restore the rank that branches destroy (Section~\ref{sec:skip}),
then perhaps having a skip connection around one matrix per branch would be
preferable. In this section, we consider a form of rank collapse in the forward
residual stream at initialization that the two-matrix structure prevents. The
mean of the residual stream over inputs is a running sum: each layer's branch
adds one term, $\mu_\ell = m_1 + \cdots + m_\ell$, where $m_k$ is the mean of
layer $k$'s branch output. When this sum grows large, it creates a rank-1 mean
spike that dominates the spectrum of the representation matrix
$Z_\ell \in \mathbb{R}^{n \times d}$, whose rows are the stream $h_\ell$ for $n$ probe
inputs, and the representations of all inputs converge to this single direction. When a single uncentered branch keeps these
terms aligned across layers, the sum grows linearly with depth, and this spike
collapses the rank at moderate depth (Figure~\ref{fig:coherence}). We show that
two architectural choices prevent this collapse: a second branch matrix
decorrelates the terms, or a centered activation avoids the mean spike in the
first place. We measure the correlation between the branch means directly and
show that it equals the activation moment ratio
$c_\ell = \mathbb{E}[\sigma]^2/\mathbb{E}[\sigma^2]$, and find that a single
criterion controls whether an architecture avoids this mean spike collapse: it
must remove the aligned component of the branch means.

We first consider the residual stream after a single layer, which
\citet{Davis2025WhenDS} show contains a mean spike in the case of an uncentered
activation function. At initialization,
the branch adds a mean of predictable size to the stream: the weights are
isotropic and the Pre-Norm fixes the branch input scale, so each coordinate of
the post-activation $\sigma(Wz)$ has mean $\mathbb{E}[\sigma]$ and second moment
$\mathbb{E}[\sigma^2]$ over the Gaussian pre-activations. The mean direction
therefore carries a fraction
$c_\ell = \mathbb{E}[\sigma]^2/\mathbb{E}[\sigma^2]$ of the post-activation energy,
where energy is the squared Frobenius norm.
\citet{Davis2025WhenDS} prove that this creates a rank-1 mean spike in the
branch-output spectrum for an uncentered activation, whereas a centered
activation has zero mean and therefore no spike; $s_\sigma = 1/c_\ell$ is an upper
bound for the stable rank of the branch output, the ratio of total to
top-direction energy. Our first contribution in this section is to show the bound is
tight at initialization across a set of common pointwise activations:
Table~\ref{tab:activations} lists the constants, and the measured stable rank
matches $s_\sigma$ to within $6\%$ in every case.

Our main contribution in this section is to show what happens to this mean
spike across layers: the spikes of successive layers can align and reinforce
one another, or point in unrelated directions and average out. As such, this
alignment controls the rate at which the mean spike grows across depth, and
therefore the rate of rank collapse: the sum of aligned means grows linearly
with $\ell$, whereas the sum of uncorrelated means is a random walk and grows
only as $\sqrt{\ell}$. Since centered activations avoid this mean spike entirely, we focus on the
setting with uncentered activations and consider what happens with one matrix
per branch versus two matrices per branch. \textbf{We find that with one matrix per branch, the correlation between the
branch means of any two layers equals the activation moment ratio $c_\ell$; with two
matrices per branch, the second matrix decorrelates the branch means and
prevents the rank collapse (Figure~\ref{fig:coherence}a; all layer pairs in
Figure~\ref{fig:coherence-heatmaps}).} Across all layer
pairs, the measured correlation is $0.320$ for ReLU ($1/\pi \approx 0.318$) and
$0.188$ for GELU (predicted $c_\ell = 0.187$; Table~\ref{tab:activations}) with
one matrix, and zero within noise with
two (Appendix~\ref{app:coherence}). To
see why a geometric quantity relating two different layers should equal a
scalar property of the activation function alone, consider where each layer's
mean comes from: with a single matrix the
activation writes directly into the stream, and an uncentered activation has
positive mean output in every coordinate, so every layer's branch mean contains
the same component along the all-ones direction, regardless of that layer's
weights. Each mean is this shared component, which carries a $c_\ell$ fraction of
its energy, plus a weight-dependent remainder that is independent across
layers; when the means of two layers are correlated, the remainders cancel and
only the shared component survives, giving exactly $c_\ell$. A second matrix
removes the shared component: each layer's $W_2$ rotates its all-ones part to a
different random direction, so the means of different layers share nothing and
their correlation is zero. When we measure the mean magnitude across depth, we see the linear mean spike
growth for the single-matrix branch (fitted exponent $0.99$) and approximately
$\sqrt{\ell}$ growth for the two-matrix branch (fitted exponent $0.56$); as
expected, the centered activation's mean is within noise of zero
(Figure~\ref{fig:coherence}b).

\begin{table}[t]
  \centering
  \caption{\textbf{Branch coherence $c_\ell$ determines the per-layer mean-spike rank,
  and the bound is tight at initialization.} For each pointwise activation
  $\sigma$, the coherence constant $c_\ell=\mathbb{E}[\sigma]^2/\mathbb{E}[\sigma^2]$
  (standard Gaussian pre-activation) determines the per-layer stable-rank bound
  $s_\sigma=1/c_\ell$ of \citet{Davis2025WhenDS}, who compute the ReLU case
  ($s_\sigma = \pi$). Uncentered activations
  ($c_\ell>0$) carry a low-rank mean spike; the measured post-activation stable
  rank $\mathrm{st}(\sigma(Wh))$ at initialization ($d{=}256$, mean over 3
  seeds) matches $s_\sigma$ to within $6\%$. Centered activations have
  $c_\ell{=}0$: no mean spike ($s_\sigma{=}\infty$) and the stable rank stays high.}
  \label{tab:activations}
  \begin{tabular}{lcrrr}
    \toprule
    Activation $\sigma$ & Centered? &
    \multicolumn{1}{c}{\begin{tabular}{@{}c@{}}Coherence\\ $c_\ell=\mathbb{E}[\sigma]^2/\mathbb{E}[\sigma^2]$\end{tabular}} &
    \multicolumn{1}{c}{\begin{tabular}{@{}c@{}}Spike bound\\ $s_\sigma=1/c_\ell$\end{tabular}} &
    \multicolumn{1}{c}{\begin{tabular}{@{}c@{}}Measured\\ stable rank\end{tabular}} \\
    \midrule
    abs   & \ding{55} & $2/\pi \approx 0.637$ & $\pi/2 \approx 1.57$ & $1.56$ \\
    ReLU  & \ding{55} & $1/\pi \approx 0.318$ & $\pi \approx 3.14$   & $3.09$ \\
    GELU  & \ding{55} & $0.187$               & $5.35$               & $5.16$ \\
    SiLU  & \ding{55} & $0.120$               & $8.33$               & $7.86$ \\
    \midrule
    tanh  & \ding{51} & $0\phantom{.000}$ & $\infty\phantom{.00}$ & $50.66$ \\
    \bottomrule
  \end{tabular}
\end{table}

The consequence appears in the spectrum of the representation: by the final
layer, the top singular value of the single-matrix branch carries essentially
all of the energy ($1.00$), whereas the two-matrix branch has a bounded spike
($0.70$) and the centered activation has no spike at all ($0.08$;
Figure~\ref{fig:coherence}c). What matters for the rank is the energy that
remains outside the spike: the two-matrix branch's remaining $0.30$ spreads
over hundreds of directions, and its effective rank stays near half the width,
whereas the single-matrix branch retains essentially nothing and its rank
collapses toward zero (Figure~\ref{fig:coherence}d). In
Appendix~\ref{app:coherence}, we show that this rank collapse is similar across
model widths. In Section~\ref{sec:skip}, we show that skip connections
preserve the Jacobian rank through their identity terms in the backward pass;
the coherent mean spike instead collapses the rank of the forward residual
representation, or the skip pathway itself. This is why the feedforward block
needs its second matrix: the projection after the activation function
decorrelates the means before they are added to the residual stream, preventing
this mean spike mechanism for rank collapse.

We note that the backward pass is also a residual stream: the gradient of the
loss with respect to each layer's representation accumulates as
$\delta_{\ell-1} = \delta_\ell + \beta\,J_\ell^\top\delta_\ell$. One might expect
a similar spike in the across-input mean of these gradients, but at
initialization it does not form: across one matrix, two matrices, and a centered
activation, the per-layer contributions are uncorrelated between layers and the
gradient stream keeps its rank (Appendix~\ref{app:coherence}). The forward spike
comes from the activation's mean, a fixed direction added at every layer; the
backward contribution instead applies the activation's derivative as a mask and
the transposed weights as a rotation, so no shared direction builds up. This
decorrelation needs the layers' weights to be independent, so whether it
persists once training couples them is open.

We next consider how normalization layers affect this mean spike, as it would
seem that a normalization operation could subtract this mean and prevent the
representation collapse. We note, however, that this mean is created by the
activation function, so a Pre-Norm operation that centers the branch input does
not remove the mean; it would require a normalization operation in the branch
after the activation function, before the branch is added to the residual
stream. We do find that mean subtraction in the Pre-Norm placement slows the
rank collapse, as it breaks a feedback loop in which the stream's accumulated
mean re-enters the branch and reinforces itself. But the collapse still occurs
because the activation re-creates the mean at the branch output regardless of
its input. A normalization placed after the activation on the branch output can
subtract the mean to remove the spike and preserve the rank. It is the mean
subtraction in particular that is important here; a rescaling operation like
RMSNorm leaves the rank collapse unchanged (Appendix~\ref{app:coherence}). Standard
transformer blocks do not need this extra normalization, however, as the second
matrix already removes the aligned component of the branch means.

The mean spike could impair training in two ways: through the magnitude it adds
to the residual stream, or through the rank it removes from the representation.
We train the networks on CIFAR without normalization, and the single-matrix
architecture fails to train by about depth 50 whereas the two-matrix
architecture fails by about depth 100 (Figure~\ref{fig:coherence-training}). We
measure the residual magnitude at initialization and find it grows exponentially
with depth, twice as fast for the single-matrix architecture, suggesting its
failure at half the depth is due to the faster magnitude growth. Adding
Pre-RMSNorm to control the magnitude restores training for both architectures,
which reach similar best accuracy over the learning rate sweep (within a point
at every depth up to 1000), even though the single-matrix representation has
still collapsed to rank around two at initialization. Since CIFAR is a relatively easy task, whether the rank
collapse impairs training at larger scale or on harder tasks remains open.

The quantity that determines the representation spectrum at depth is the
cross-layer correlation of the branch means. This section showed it equals a
scalar property of the activation, $c_\ell = \mathbb{E}[\sigma]^2/\mathbb{E}[\sigma^2]$.
The per-layer stable-rank bound in \citep{Davis2025WhenDS} predicts the
conditioning of the gradient, but no per-layer quantity can predict whether the
rank survives with depth: a per-layer bound sees the singular values of one
layer and not their directions, and the single-matrix and two-matrix branches
carry the same per-layer spike and the same per-layer stable rank, diverging
only in whether those spikes align across layers. The cross-layer correlation
is a single scalar, cheap to measure, and it determines whether the branch
means accumulate coherently or average out across depth.

\section{Width Expansion and the Full-Rank Branch Threshold}
\label{sec:width}

\begin{figure*}[t]
  \centering
  \includegraphics[width=0.325\textwidth]{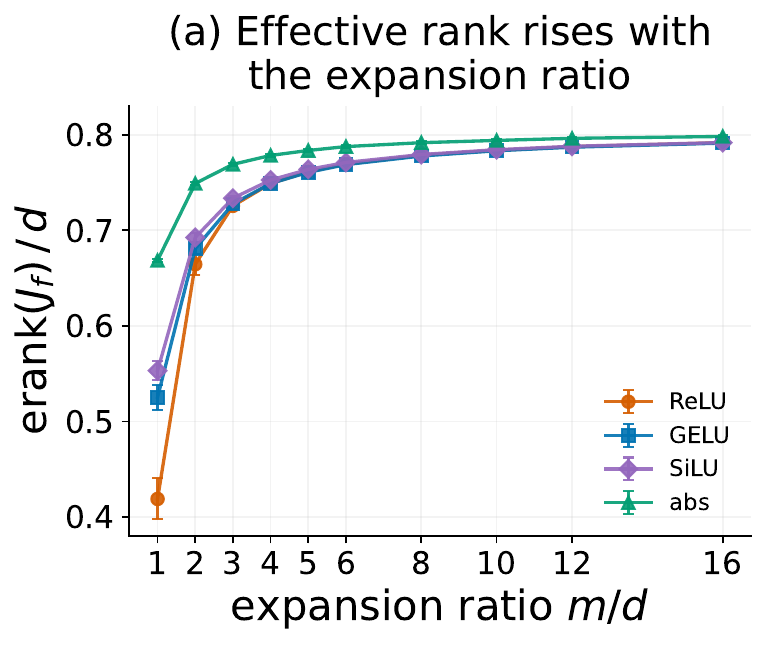}\hfill
  \includegraphics[width=0.325\textwidth]{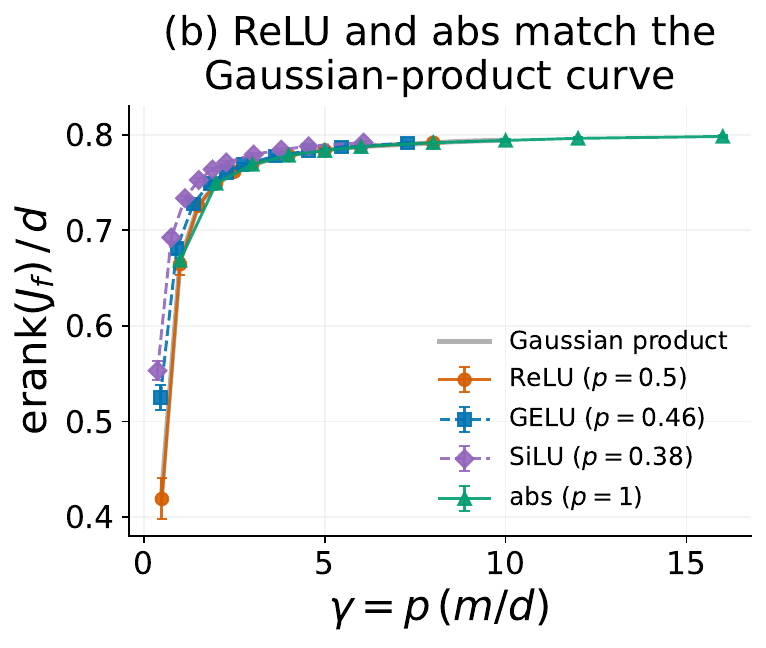}\hfill
  \includegraphics[width=0.325\textwidth]{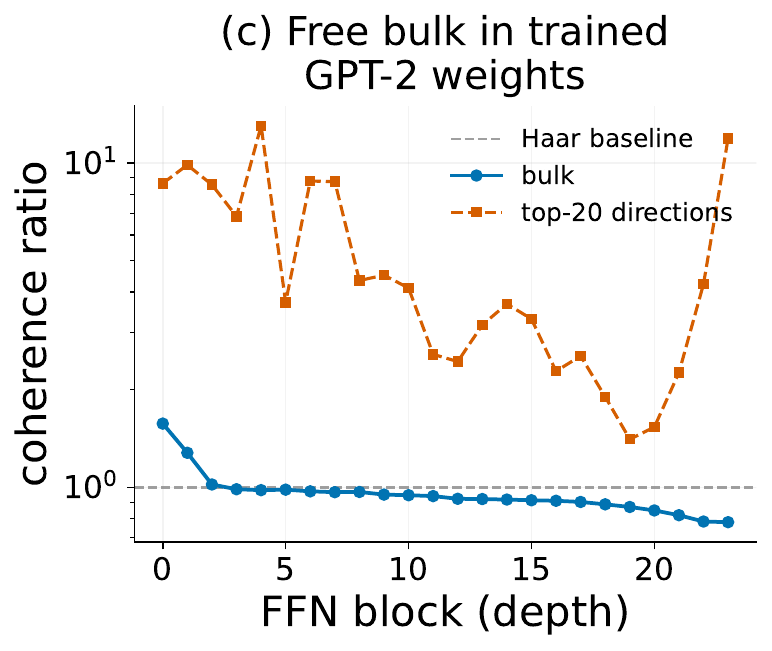}
  \caption{\textbf{The branch Jacobian's effective rank increases with the FFN
  expansion ratio, and a Gaussian-product law predicts it.}
  \textbf{(a)} The normalized effective rank $\mathrm{erank}(J_f)/d$ increases
  with the expansion ratio $m/d$ for ReLU, GELU, SiLU, and abs, fastest for abs
  ($p=1$); GELU and SiLU reach a higher effective rank than ReLU at small
  expansion ratios despite their smaller $p$.
  \textbf{(b)} Plotted against $\gamma = p\,(m/d)$, ReLU and abs (with $p$ the
  survival rate) fall on the effective-rank curve of a product of two Gaussian
  matrices, while GELU and SiLU (with $p = \E[\sigma'^2]$) approach it from above.
  \textbf{(c)} The prediction assumes $W_{\mathrm{up}}$ and $W_{\mathrm{down}}$
  are asymptotically free; in pretrained GPT-2 Medium their singular-vector
  coherence, averaged over all directions, stays close to the Haar baseline of $1$
  across the $24$ FFN blocks
  (mean $0.96$), so the assumption holds in a trained model. A few extreme
  directions (the top and bottom $20$) are more aligned, but lie outside the
  bulk that determines the effective rank. $d = 256$, $40$ seeds in (a,b); the
  Gaussian-product curve in (b) uses $20$ seeds and spans $\gamma \le 10$.}
  \label{fig:width-cond}
\end{figure*}

\begin{figure*}[t]
  \centering
  \includegraphics[width=0.86\textwidth]{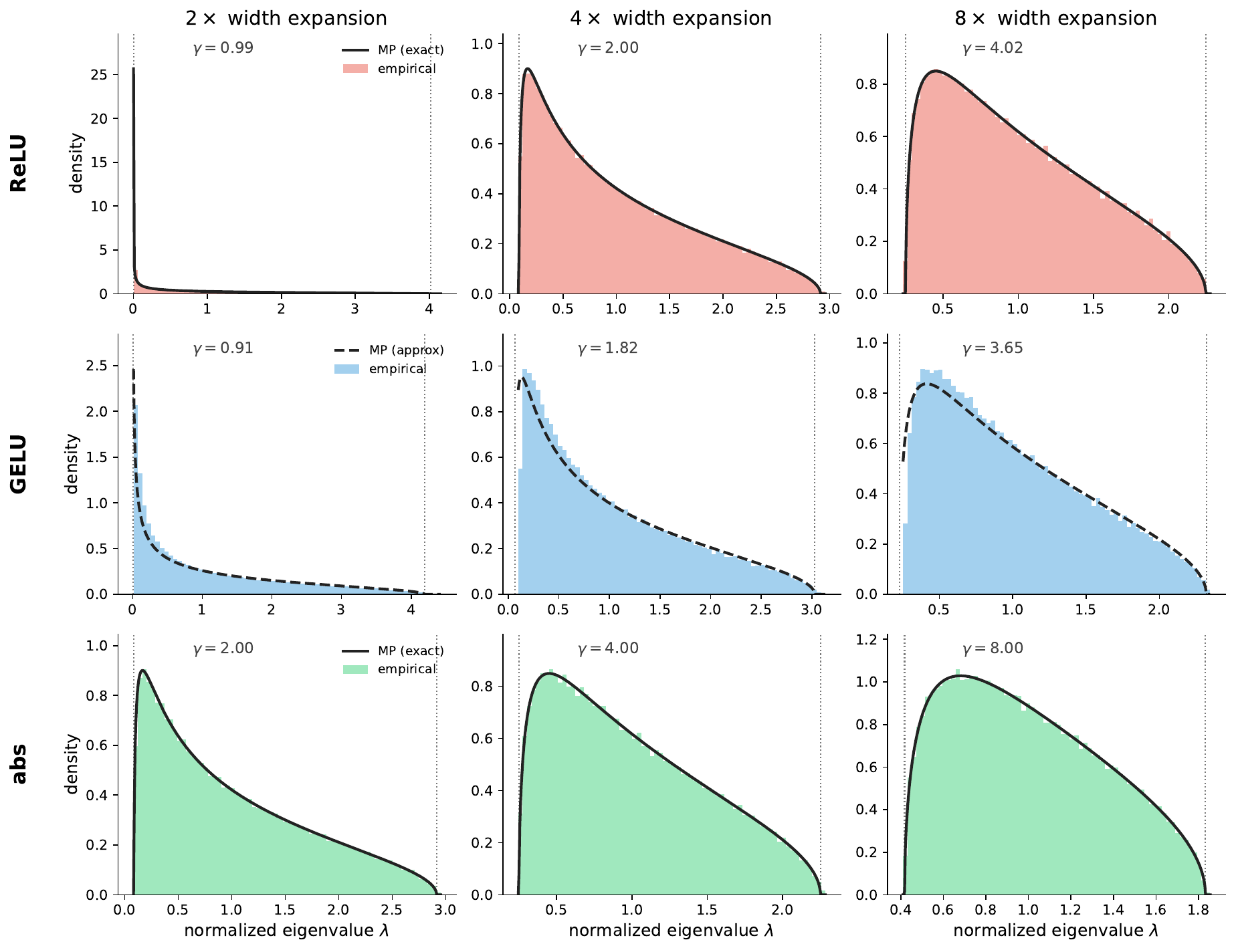}
  \caption{\textbf{The masked up-projection follows the Marchenko--Pastur law,
  and its conditioning improves as $m/d$ rises above $1/p$.} Eigenvalue
  histogram (filled) of the Gram matrix
  $G = \tfrac{1}{n_{\mathrm{eff}}}(D\,W_{\mathrm{up}})^{\top}(D\,W_{\mathrm{up}})$
  of the masked up-projection $D\,W_{\mathrm{up}}$ against the Marchenko--Pastur
  density (line), for ReLU, GELU, and abs (rows) at width expansions
  $m/d \in \{2, 4, 8\}$ (columns). The mask
  $D = \diag(\sigma'(W_{\mathrm{up}}x))$ keeps a fraction $p$ of the $m$ rows, so
  the survivor count is $n_{\mathrm{eff}} = \sum_i \sigma'(z_i)^2$ and the aspect
  ratio is $\gamma = n_{\mathrm{eff}}/d$. The eigenvalues of $G$ then lie on the
  support $[(1-1/\sqrt\gamma)^2, (1+1/\sqrt\gamma)^2]$ (dotted edges). At
  $\gamma = 1$ ($m/d = 1/p$, or $2\times$ for ReLU) the lower edge of the
  spectrum is zero and $D\,W_{\mathrm{up}}$ is badly conditioned; for
  $\gamma > 1$ the spectrum separates from zero. Since $W_{\mathrm{down}}$
  is full rank, this submatrix sets $\rank J_f$. Exact for ReLU and abs (solid),
  approximate for GELU (dashed). Random Gaussian $W_{\mathrm{up}}$ with
  unit-variance entries, $d = 512$, $40$ seeds. See Appendix~\ref{app:mp} for SiLU.}
  \label{fig:width-spectrum}
\end{figure*}

The previous section illustrates why the Transformer feedforward block uses two
weight matrices inside every residual branch. In practice, these two matrices
have a particular structure: the first matrix expands the width before the
activation function is applied, then the second matrix contracts back to the
original width. In this section, we ask how this width expansion ratio between
the feedforward dimension and model dimension affects the gradient along the
residual branch. We write the branch as
$f(x) = W_{\mathrm{down}}\,\sigma(W_{\mathrm{up}} x)$, with up-projection
$W_{\mathrm{up}} \in \R^{m \times d}$ and down-projection
$W_{\mathrm{down}} \in \R^{d \times m}$. We show that the width expansion factor
$m/d$ controls the rank of the branch Jacobian $J_f$ through the
ratio $\gamma = p(\sigma)\,(m/d)$: the activation mask keeps a fraction
$p(\sigma)$ of the rows in the up-projection matrix, and the surviving submatrix
(and hence $J_f$) becomes full rank only once $m/d$ reaches $1/p(\sigma)$. At this
threshold the submatrix is full rank but badly conditioned, as its smallest
singular value is near zero (Figure~\ref{fig:width-spectrum}). For ReLU, the
full-rank threshold is a $2\times$ expansion, but the $4\times$ expansion used in
practice \citep{vaswani2017attention} is much better conditioned.

We can differentiate the branch to write its Jacobian as follows, where $D$ is a
diagonal mask that keeps only the hidden units with nonzero derivative:
\begin{equation}
  J_f = W_{\mathrm{down}}\, D\, W_{\mathrm{up}},
  \qquad D = \diag\!\big(\sigma'(W_{\mathrm{up}} x)\big).
  \label{eq:jac-factor}
\end{equation}
Let $p(\sigma)$ denote the expected fraction of the $m$ rows that survive the
mask $D$; at initialization
$p(\sigma) = \Pr[\sigma'(z)\neq 0]$ for $z\sim\mathcal{N}(0,1)$, which is one half for
ReLU and coincides with the constant $p = \mathbb{E}[\sigma'(z)^2]$ in
Section~\ref{sec:skip} for masking activations. The Gaussian submatrix formed by the $pm$ of the $m$ rows that survive the
mask has Gram matrix $\tfrac{1}{pm}(D\,W_{\mathrm{up}})^{\top}(D\,W_{\mathrm{up}})$,
whose eigenvalues follow the Marchenko--Pastur law with aspect ratio
$\gamma = pm/d$, in the proportional limit $d, m \to \infty$ (Appendix~\ref{app:mp}).
Because $W_{\mathrm{down}}$ is full-rank and independent of these rows at
initialization, it does not reduce the rank of the submatrix, so
$\rank J_f = \rank(D\,W_{\mathrm{up}})$.
The $pm$ surviving rows span the $d$-dimensional input space only when $pm \ge d$,
that is $\gamma \ge 1$. Because the mask fixes the fraction $p$, widening the branch
is the only way to push $pm$ past $d$ so that $D\,W_{\mathrm{up}}$, and therefore
$J_f$, is full rank.
The Marchenko--Pastur law gives the spectrum on
$[(1-1/\sqrt\gamma)^2, (1+1/\sqrt\gamma)^2]$ (Figure~\ref{fig:width-spectrum}),
where the lower edge, the smallest eigenvalue, is zero at $\gamma = 1$ and
separates from zero as $\gamma$ increases. The measured spectra match this
prediction exactly for ReLU and abs, and approximately for GELU and SiLU.
Setting $\gamma = 1$ gives the threshold $m/d = 1/p(\sigma)$, which corresponds
to a $2\times$ width expansion for ReLU. The threshold is correspondingly higher
for the smooth activations, about $2.2\times$ for GELU and $2.6\times$ for SiLU, with
$p = \E[\sigma'^2]$ as explained below; at $m/d = 2$ in Figure~\ref{fig:width-spectrum},
their spectral lower edge has not separated from zero.

The effective rank of the branch Jacobian increases with the expansion ratio
$m/d$ for all activations (Figure~\ref{fig:width-cond}a). For activations whose derivative takes only the
values $0$ and $\pm1$ (ReLU and abs), the masked up-projection
$D\,W_{\mathrm{up}}$ is exactly Gaussian, so $J_f$ is a product of two Gaussian
matrices and we can compute its effective rank as a function of
$\gamma = p\,(m/d)$ (Figure~\ref{fig:width-cond}b). For these two activations,
$p$ is the survival rate, $\tfrac12$ and $1$. For GELU and SiLU, $\sigma'$
reweights each coordinate instead of masking it, so $p$ is instead
$\E[\sigma'^2]$, the average squared derivative ($0.46$ and $0.38$); their curves
approach the Gaussian-product curve from above (Appendix~\ref{app:mp}).
Because reweighting attenuates directions without removing them, the smooth
activations retain more rank than masking at the same $p$: at small expansion
ratios, GELU and SiLU reach a higher effective rank than ReLU despite their
smaller $p$.
The Gaussian-product law assumes $W_{\mathrm{up}}$ and $W_{\mathrm{down}}$ are
free; we measure pretrained GPT-2 and show the bulk singular-vector coherence
stays close to the Haar baseline across the 24 FFN blocks
(Figure~\ref{fig:width-cond}c).
While the residual skip preserves the rank of the propagated gradient through
its identity term, the feedforward width expansion is another architectural
factor that controls the rank of the branch Jacobian itself.

\section{Related Work}
\label{sec:related}

The literature on dynamical isometry, the condition where every singular value of the
input--output Jacobian remains near one so that rank is preserved across depth, shows
that narrow conditions on the initializations and activation functions must be
satisfied. The Jacobian is a product of weight matrices and activation derivatives, and
every factor must be a near-isometry, because the product compounds any spread in the
singular values across depth. For the weights, this requires orthogonal initialization,
since a matrix with all of its singular values equal to one is orthogonal
\citep{saxe2014exact}. For the activations, the derivative must be near-constant in
magnitude on the bulk of the preactivation distribution \citep{pennington2017resurrecting},
leaving only two categories. Nearly linear activations, with a large linear region
around the origin, attain approximate isometry but lack expressivity;
absolute-value-like activations achieve perfect isometry but collapse the forward-pass
correlations, driving distinct inputs to the same representation
\citep{murray2021activation}. In particular, ReLU has derivative zero on half of its
inputs and cannot achieve dynamical isometry. \citet{tarnowski2019dynamical} obtain
dynamical isometry in residual networks for any activation function by downscaling the
branch: this is one end of the tradeoff we study, where a skip-dominated network
preserves rank but provides less gradient signal encouraging the layers to compose
(Section~\ref{sec:skip}).

Outside these narrow conditions, \citet{feng2022rank} prove that for layers that are
smooth almost everywhere, the rank of the input--output Jacobian can only decrease with
depth, and measure this decay in pretrained ResNets, MLP-mixers, and vision
transformers. In their exact rank framework, skip connections are a complete fix,
because a residual block with a small branch loses no rank at all. The effective rank
we track instead degrades continuously through every block, and we quantify how the
branch scale, initialization, normalization placement, and width determine how much
survives.

Another line of work studies rank collapse along the token dimension, where
self-attention drives the representations of all tokens toward a common vector.
\citet{dong2021attention} prove that a network of pure self-attention layers collapses
doubly exponentially toward this rank-one state, and that skip connections can
counteract the collapse while MLPs only slow it. \citet{noci2022signal} show that when
the token representations collapse at initialization, the gradients of the queries and
keys vanish, and they prevent the collapse by scaling the branches by $1/\sqrt{L}$.
\citet{naitsaada2025mindthegap} attribute the collapse to the spectral gap of the
softmax attention matrix, and show that token representations also collapse as the
context length grows. Whether the collapse occurs depends on the surrounding
components: with LayerNorm the dynamics admit stable equilibria of every rank
\citep{wu2024attnmasksln}, and with skip connections the outcome is determined by the
spectrum of the value and projection weights \citep{dovonon2025oversmoothing}. Even
where skip connections do preserve the token rank, small weights leave the network
equivalent to a single attention layer \citep{alman2025only}. The same phenomenon
appears as over-smoothing in vision transformers and BERT
\citep{wang2022antioversmoothing, shi2022revisiting}. These results concern the token
dimension of attention layers; the rank we study is along the feature dimension, in the
gradients and representations of the feedforward blocks.
\citet{baker2024gradientrank} study gradient rank collapse in a third sense, the rank
of the weight gradients accumulated over a batch, which is bounded by the narrowest
layer width, the batch size, and the linearity of the activation.

Deep residual networks have been shown to behave like ensembles of shallow subnetworks
rather than deep networks whose layers compose. \citet{veit2016residual} decompose the
forward pass into $2^L$ paths that take either the skip connection or the residual
branch at each layer, and show that the gradient magnitude is carried mostly by the
short paths, while deleting or reordering individual layers barely changes the output
of a trained network. \citet{dherin2025residual} use a Taylor series expansion of the
forward pass in the residual branch scale, so that a deeper network corresponds to a
larger implicit ensemble, and show that controlling the magnitude growth of this
expansion by downscaling the branch enables training without normalization.
\citet{liu2026inversedepth} find that the depth scaling exponent of language model loss
matches what ensemble averaging would predict, and that the middle layers of trained
models make similar, weakly correlated updates. Through a similar path decomposition on
the input--output Jacobian, we measure how the gradient magnitude distributes over the
path orders, where higher orders pass through more rank-reducing branch operations
(Section~\ref{sec:skip}). Ensemble behavior and rank collapse
are then the two ends of a single tradeoff, and the branch-to-skip ratio determines
where a network falls between them.

The placement of normalization layers has been explored extensively as a design choice:
architectures add normalizations at the branch input and output
\citep{ding2021cogview,kim2025peri}, inside the branch
\citep{shleifer2021normformer,wang2023magneto}, on the branch alone
\citep{liu2024branchnorm}, varying across depth \citep{li2025mixln}, or in combination
\citep{zhuo2025hybridnorm}. The difficulty of training deep Post-Norm networks has been
explained through magnitudes: \citet{xiong2020layer} show that Post-Norm concentrates
large gradients near the output at initialization, motivating warmup, and
\citet{wang2024deepnet} bound the update magnitude to train thousand-layer Post-Norm
networks. From the parameterization perspective, \citet{dey2025completep} tune the depth
scaling so that hyperparameters transfer and every layer continues to learn features.
Using signal propagation analyses, \citet{noci2023shaped} modify the attention to keep
its covariance well-behaved at large depth, \citet{he2023transformersnoshortcut} train
deep transformers without skip connections or normalization, and
\citet{he2023simplifying} remove block components without losing training speed. These approaches
design for the magnitude and stability of signals and updates; we show that the
normalization placement also determines the effective rank, through the branch-to-skip
ratio across depth (Section~\ref{sec:norm-placement}). \citet{blake2024umup} prove that
any Pre-Norm network is exactly equivalent to a network with unit-scale stream whose
per-layer multipliers preserve the branch-to-skip scale ratio, and use this equivalence
to hold activations at unit variance at every depth for low-precision training. We
connect the same per-layer ratio, which decays as the Pre-Norm stream grows, to the
Jacobian rank.

\section{Discussion}
\label{sec:discussion}

Through the lens of preserving rank across depth, we have seen that the architectural
components of the Transformer feedforward block act through two different mechanisms: the
first routes much of the gradient around the residual branch through the skip
connections, whereas the second increases the rank preserved in the residual branch
itself. The two mechanisms act on three distinct rank objects: the first preserves the
input--output Jacobian, whereas the second preserves the residual representation and the branch Jacobian.

The first mechanism works around a rank loss that is intrinsic to deep networks:
outside the narrow conditions of dynamical isometry \citep{saxe2014exact,
pennington2017resurrecting}, every matrix multiplication and activation reduces the
rank, and the losses compound with depth. Adding skip connections to MLP layers does
not avoid this intrinsic rank loss, since the branch operations remain rank-reducing;
instead, the skip connections simply restore the rank periodically by carrying the loss
gradient around the branch. Our analysis shows that the more of the gradient that travels
through the skip rather than the branch, the more rank survives across depth. A small
initialization or small branch scale both reduce the branch-to-skip ratio, routing more
of the gradient around the rank-reducing operations.

Normalization layers are usually understood as a way to keep the network magnitudes
stable across depth; our analysis shows they also govern rank, by the same routing
mechanism as the branch scale. A normalization placed along the residual branch, at the
branch input or output, never rescales the residual stream, so the stream grows
with depth while each branch adds an increment of fixed scale. The branch then contributes less to the
stream in deeper layers, just as a smaller branch scale would, even with residual branch
scale $\beta = 1$.
A post-residual rescaling does the opposite, renormalizing the stream at every block and
holding its scale fixed, so the branch-to-skip ratio stays constant and the rank
collapses with depth. The rank collapse is driven by the branch-to-skip ratio, not the
projection terms that appear in the normalization gradient. This perspective on the first mechanism, the
routing of the gradient between the branch and the skip, unifies the remedies proposed
to prevent rank collapse \citep{noci2022signal}, the conditions under which residual
networks achieve dynamical isometry \citep{tarnowski2019dynamical}, and depth-scaling
prescriptions motivated by other goals \citep{yang2023depth, bordelon2023depthwise}:
each relies on a small branch-to-skip ratio, whether from a
small branch scale or a small initialization.

The second mechanism increases the rank the branch itself preserves, through the
two-matrix structure of the feedforward block where the first matrix expands the hidden
dimension, the activation is applied in the expanded space, and the second matrix
contracts back to the original dimension. The two-matrix structure alone, even without
the width expansion, prevents the branch means across layers from accumulating
coherently: the down-projection matrix decorrelates the branch means, so the mean spike
stays bounded and the representation keeps its rank. The width expansion structure
plays another important role: intuitively, the up-projection mixes the input directions,
so that the elementwise activation that zeros or attenuates some of these mixed directions
loses part of many input directions rather than collapsing any single one entirely. This
allows the down-projection to recombine, in the original dimension, the directions that
survive the activation; the wider the expansion, the higher the branch Jacobian's
effective rank (Section~\ref{sec:width}).

These two mechanisms illustrate a fundamental tradeoff in architecture design for deep residual networks:
even when the residual branch is designed to preserve as much rank as possible when applying matrix
multiplications and activation functions, the gradient at each layer flows with some ratio through the
rank-reducing operations in the residual branch compared to the rank-preserving identity in the skip connections.
If the only objective were to preserve rank in the input--output Jacobian, the trivial solution would be to have
all of the gradient flow through the skip connections; however, this makes the network act like an ensemble
of layers rather than a deep neural network whose layers learn to compose. This tradeoff reframes the ensemble
behavior that prior work observes in deep residual networks \citep{veit2016residual, dherin2025residual}: the
gradient magnitude concentrates on the short paths through few rank-reducing operations
(Section~\ref{sec:skip}), so the same routing that preserves rank is what makes the
network behave like an ensemble; under this perspective, the ensemble regime is not
simply a pathology but a real alternative to rank collapse. The second mechanism escapes this tradeoff,
preserving rank inside the branch rather than routing the gradient around it, but pays
for the rank with additional parameters. The feedforward block design then navigates a
three-way tradeoff among rank collapse, ensemble-like behavior, and parameter count.

Our perspective on the path decomposition of the input--output Jacobian shows us that the
branch-to-skip ratio unifies how the architectural choices, including the initialization
scale, residual branch scale, and normalization layers, affect this tradeoff. The choice of the initialization scale and residual branch scale directly control
this ratio, but as shown in Section~\ref{sec:norm-placement}, the normalization layers
\emph{also} control the branch-to-skip ratio and therefore indirectly also control this
tradeoff. In practice, many frontier models keep the residual branch scale $\beta = 1$
rather than scaling the branch down (Section~\ref{sec:skip}); this suggests that shifting
the gradient too much into the skip connections has empirical disadvantages, perhaps in
the network's ability to compose its layers. \citet{liu2026inversedepth} find that
language model loss improves with depth at the exponent predicted for an ensemble of
shallow subnetworks, rather than the faster exponent expected when the layers compose,
suggesting that trained language models operate near the ensemble regime. This is
consistent with the mechanism in Section~\ref{sec:norm-placement}: even at $\beta = 1$,
Pre-Norm lets the effective branch-to-skip ratio decay with depth, so the deeper layers
receive increasingly skip-dominated gradients. On synthetic tasks that require multi-step reasoning,
however, deeper transformers outperform shallower but wider models with equal or larger
parameter counts \citep{ye2024physics2p1}. Where a network should sit on this tradeoff
is then likely task-dependent: next-token prediction appears well served by the ensemble
regime, whereas reasoning requires the layers to compose.
\section{Limitations and Future Work}
\label{sec:limitations}

Our analysis of rank focuses on the feedforward block at initialization, leaving
questions about how attention layers and the training process affect the gradient and
representation spectra. In principle, mixing across the sequence dimension could offset
rank loss by combining representations across positions; in practice, prior work shows
that self-attention layers can collapse the token representations toward rank one
\citep{dong2021attention}. Tracking how these spectra evolve during training, and how the
optimizer affects them, is a natural extension for future work. We show that the effective
rank of the input--output Jacobian predicts CIFAR-10 trainability in
Section~\ref{sec:norm-placement}, but as noted there, CIFAR-10 is an easy task that
provides only a coarse signal of trainability.
A more challenging task, in particular one that requires high-rank representations or
sequential composition between layers, could show how the effective rank in the gradient
and representation spectra matters in tasks where depth is more genuinely required.

Our derivations explain how each component preserves rank, but leave several specific
questions open. For normalization layers, our scaling heuristic argues that
$\beta\alpha^{M}\sqrt{L}$ controls the rank; we measure the shape of the rank curve
empirically. For the width expansion, our threshold $m/d = 1/p$ is exact
only for activations whose derivative is $0$ or $\pm 1$; our results are approximate for
other pointwise activations and we leave gated activations as an open question. Both
results could also be applied directly, to predict how deep a normalization placement can
scale before its rank collapses and to choose the width-expansion factor that keeps the
branch Jacobian full rank for a given activation.

Any neural network architecture that scales depth faces intrinsic rank attenuation for
gradients passing through long paths of rank-reducing operations; in deep residual
networks, this appears as a tradeoff between rank collapse and ensemble-like behavior. This may induce fundamental
limits on the depth-scaling exponents that each architecture family can attain, or
possibly universal limits on scaling exponents for neural networks as a whole. At the same
time, there may be advantages to mild or moderate amounts of rank loss: limitations on the
rank of gradients or representations may provide implicit regularization and improve
generalization \citep{arora2019implicit, Huh2021TheLS}, or act as a useful form of
compression \citep{galanti2025sgd}.

Our path decomposition arguments about gradient signals give some intuition about why
skip-dominated gradients may encourage ensemble behavior, but the training dynamics of
layer composition are not well-understood. Intuitively, the gradient component through the
residual branch tells the feedforward blocks how they should update so that they can
cooperatively reduce the loss; in other words, the longer the path, the more the gradient
encourages the layers to compose. At the extreme, a network that routes all of its gradient through the skip is a
pure ensemble whose layers cannot learn to compose. However, even gradient signal through paths with only
several MLP layers may still be sufficient for many layers to gradually learn to compose:
if adjacent layers learn to cooperate and progressively pull in their neighbors, deep
composition could assemble without gradients ever traversing particularly long paths.
Moreover, our path decompositions consider the gradient magnitudes, but gradient
signal with small magnitude need not be small in its effect on the function a network
learns. Indeed, \citet{zhang2024initialization} find that transformers initialized
below the typical scale, where the residual branches are weakest, are the ones that
converge to compositional rather than memorized solutions, an effect they hypothesize
stems from the low-complexity bias of small initializations. Characterizing the relationship
between gradient path lengths and how networks actually learn to compose layers is a
natural question for future work.

Finally, future work building on our analysis could use spectral properties of the
gradient and representation not just to understand how existing architectures work, but to
design new architectures and predict how they scale and contend with these fundamental
tradeoffs. The tradeoffs we describe likely take different forms in other architecture
families, such as recurrent networks and state-space models. Understanding how each
component affects these spectra is a step toward combining them into new architectures by
design rather than by trial and error.

\section{Conclusion}
\label{sec:conclusion}

We have analyzed the modern feedforward block component by component and shown that each
one helps preserve the rank of the signals the network propagates across depth: the skip
connection routes the gradient around the rank-reducing branch, the placement of
normalization sets the branch-to-skip ratio across depth, and with it how much rank
survives, the second matrix keeps the representation from
collapsing onto a coherent spike, and the width expansion keeps the branch Jacobian full
rank. Each of these components was originally motivated by a different purpose --- skip
connections to prevent vanishing gradients, normalization to control representation
magnitudes, the second matrix and width expansion to add capacity --- but seen through the
lens of rank preservation, they work together to keep the gradient and representation
ranks from collapsing with depth. Our analysis is a step toward architecture design that
navigates the tradeoff between rank preservation and layer composition as a first
principle, and toward understanding fundamental
limitations in scaling neural networks.

\clearpage
\bibliographystyle{plainnat}
\bibliography{../references}

\clearpage
\section*{Acknowledgements}
We thank Leslie Pack Kaelbling for valuable technical discussions. We are grateful
to the Google TPU Research Cloud (TRC) program, MIT CSAIL, Phillip Isola, and
Kaiming He for providing the compute resources that made this work possible.

\appendix
\section{Skip Connections and the Path Decomposition: Derivations}
\label{app:skip}

This appendix collects the derivations behind Section~\ref{sec:skip}: the path
decomposition of the input--output Jacobian (Eq.~\ref{eq:path-decomp}), the
per-order magnitudes in Figure~\ref{fig:paths}, the recurrence and numerical check
behind them, and the argument that the skip must be the identity.

\subsection{The path decomposition}

The chain rule turns the residual recurrence $h_\ell = h_{\ell-1}+\beta f(h_{\ell-1})$ into
an ordered product over the $L$ blocks,
\begin{equation}
  \frac{\partial h_L}{\partial h_0}
    = (I+\beta J_L)(I+\beta J_{L-1})\cdots(I+\beta J_1),
  \qquad J_\ell := \frac{\partial f}{\partial h}\Big|_{h_{\ell-1}}.
  \label{eq:app-product}
\end{equation}
Each $J_\ell$ is the branch Jacobian at the forward activation $h_{\ell-1}$, so the
product is evaluated along one forward pass and holds for any differentiable
branch.

Distributing the product gives one term per subset $A\subseteq\{1,\dots,L\}$: each
factor contributes either $I$ or $\beta J_\ell$, and $A$ records which blocks
contribute the branch. This expands the Jacobian into $2^L$ paths.

A subset of size $|A|=k$ carries a factor $\beta^k$ and a product of $k$ branch
Jacobians. Grouping the $2^L$ paths by $k$ defines the order-$k$ sum
\begin{equation}
  S_k := \sum_{1 \le \ell_1 < \cdots < \ell_k \le L} J_{\ell_k} \cdots J_{\ell_1},
  \qquad \frac{\partial h_L}{\partial h_0} = \sum_{k=0}^{L} \beta^k S_k,
  \label{eq:app-Sk}
\end{equation}
which is Eq.~\ref{eq:path-decomp}. The order $k$ counts the feedforward blocks a path
passes through, from $S_0=I$ (the pure skip) to $S_L=J_L\cdots J_1$ (every branch).
Each $S_k$ multiplies $k$ branch Jacobians, so it inherits the rank deficiency of
a $k$-fold branch composition, weighted by $\beta^k$.

The order-$k$ component $\beta^k S_k$ is the part of the Jacobian built from paths
through exactly $k$ feedforward blocks. We measure its size by the Frobenius norm
$\mu_k := \lVert\beta^k S_k\rVert_F = \sqrt{\sum_i \sigma_i^2}$. The components sum
to the Jacobian, $\sum_k \beta^k S_k = \partial h_L/\partial h_0$, but their norms
do not, so each $\mu_k$ reports one order on its own. Normalizing the $\mu_k$ to
sum to one gives the profile in Figure~\ref{fig:paths}, whose peak (the order of
largest magnitude) sits at order three for $\beta=1$ and at the order-0 skip for
$\beta=1/\sqrt L$ and $\beta=1/L$.

To efficiently compute the order-$k$ sum with $\binom{L}{k}$ terms, we use a
dynamic programming algorithm over the blocks. Let $S_k^{(m)}$ be the sum of
order-$k$ paths through the first $m$ blocks, so $S_k = S_k^{(L)}$. An order-$k$
path through the first $m$ blocks either takes the skip at block $m$, leaving an
order-$k$ path through the first $m-1$ blocks, or takes the branch, extending an
order-$(k-1)$ path through the first $m-1$ blocks by the factor $J_m$. Block $m$
has the highest index of the first $m$ blocks, so the descending-order product
puts $J_m$ on the left:
\begin{equation}
  S_k^{(m)} = S_k^{(m-1)} + J_m\, S_{k-1}^{(m-1)},
  \qquad S_0^{(m)} = I, \quad S_k^{(0)} = 0 \ \ (k \ge 1).
  \label{eq:app-recurrence}
\end{equation}
Filling the table for $m=1,\dots,L$ and $k=0,\dots,L$ computes every $S_k$ in
$O(L^2)$ matrix multiplications, one per $(m,k)$ cell.

For numerical verification of the recurrence, we draw a random Gaussian input and
random weights, compute the Jacobian both by the reconstruction
$\sum_k \beta^k S_k$ and by automatic differentiation of the forward pass, and
confirm the two agree to a relative Frobenius error of order $10^{-15}$. Because the
masses $\mu_k$ span many orders of magnitude, we store each $S_k$ as a
unit-Frobenius matrix and a log-norm scalar and accumulate in log-sum-exp form.
Figure~\ref{fig:paths} reports the $\mu_k$ and their normalized profile, averaged over 8 seeds, where each
seed has its own weight initialization and random Gaussian input; for the
initialization-scale panels, the draws are independent at every $(\beta, \alpha)$
grid point.

\subsection{Why the skip must be the identity under Pre-Norm}

We reproduce the argument of \citet{he2016identity} that the skip connection
should be the identity, through the path decomposition above. Consider a residual
network where the skip is scaled by a constant $\lambda$, so that each block is
$h_\ell = \lambda h_{\ell-1} + \beta f(h_{\ell-1})$ and the Jacobian factor becomes
$\lambda I + \beta J_\ell$:
\begin{equation}
  \frac{\partial h_L}{\partial h_0}
    = (\lambda I + \beta J_L)(\lambda I + \beta J_{L-1})\cdots(\lambda I + \beta J_1).
  \label{eq:app-scaled-skip}
\end{equation}
This argument needs the residual stream to pass through each block untouched, which
is satisfied by Pre-Norm (but not Post-Norm). Distributing as before, each factor
contributes either $\lambda I$ or $\beta J_\ell$,
so a path of order $k$ carries $\lambda^{L-k}\beta^k$, and the order-0 path that
stays on the skip at every block is $\lambda^L I$. This skip path passes through no
weights, so its scale $\lambda^L$ is independent of the network's weights, at
initialization or anywhere in training. The same factor governs both passes: from
block $\ell$, the skip reaches the output through $L-\ell$ scaled steps, contributing
$\lambda^{L-\ell} h_\ell$ to the forward signal $h_L$ and $\lambda^{L-\ell} I$ to the
backward Jacobian $\partial h_L/\partial h_\ell$. Because $\lambda^{L-\ell}$ is
exponential in depth, any $\lambda \ne 1$ makes the skip path explode
($\lambda>1$) or vanish ($\lambda<1$), and $\beta$ cannot correct it because the
skip path carries no factor of $\beta$. Only $\lambda=1$ holds the skip
contribution at $O(1)$ across depth, which is why the standard residual block
fixes the skip to the identity and leaves $\beta$ as the only free scale.

\subsection{Magnitude growth}

Squaring the block $h_\ell = h_{\ell-1} + \beta f(h_{\ell-1})$ gives the magnitude
recurrence
\begin{equation}
  \lVert h_\ell\rVert^2 = \lVert h_{\ell-1}\rVert^2
    + 2\beta\,\langle h_{\ell-1}, f(h_{\ell-1})\rangle
    + \beta^2\lVert f(h_{\ell-1})\rVert^2.
  \label{eq:app-magnitude}
\end{equation}
Let $g_\ell = \lVert f(h_{\ell-1})\rVert / \lVert h_{\ell-1}\rVert$ be the branch gain and
$\rho_\ell = \langle h_{\ell-1}, f(h_{\ell-1})\rangle / (\lVert h_{\ell-1}\rVert\,\lVert f(h_{\ell-1})\rVert)$
the alignment between the branch output and the stream. Then
\begin{equation}
  \lVert h_\ell\rVert^2 = \lVert h_{\ell-1}\rVert^2\,(1 + 2\beta \rho_\ell g_\ell + \beta^2 g_\ell^2).
  \label{eq:app-magnitude-factored}
\end{equation}
At initialization the two-matrix branch output is roughly uncorrelated with the stream, so
$\rho_\ell \approx 0$ and the factor is $1 + \beta^2 g_\ell^2 \ge 1$: the magnitude grows at
every block, fastest at $\beta = 1$ and not at all as $\beta \to 0$. Whether the
growth is additive or multiplicative depends on the gain. Without normalization the
branch output scales with its input, $g_\ell = O(1)$, and the factor compounds to
exponential growth in depth. A normalization on the branch input fixes the branch
output scale, so $g_\ell = \lVert f\rVert/\lVert h_{\ell-1}\rVert$ shrinks as the stream
grows; the increment $\beta^2\lVert f\rVert^2$ is then roughly constant and the
stream grows additively, $\lVert h_L\rVert^2 \approx \lVert h_0\rVert^2
+ \beta^2\sum_\ell \lVert f\rVert^2 = O(L)$.

\subsection{The branch-scale symmetry}
\label{app:skip-symmetry}

We show that for residual networks with positively homogeneous activations and biasless
branches, networks with the same $\beta\alpha^{M}$ compute the same function at
initialization. Write the branch weights as $W_i = \alpha\bar{W}_i$, for arbitrary matrices
$\bar{W}_i$. Positive homogeneity gives $\sigma(\alpha z) = \alpha\,\sigma(z)$ for
$\alpha > 0$; applying this identity at each activation, together with the
linearity of the matrices, factors the weight scale out of the branch,
\begin{equation}
  \beta f(x;\, \alpha\bar{W}_1, \ldots, \alpha\bar{W}_M)
  = \beta\alpha^{M} f(x;\, \bar{W}_1, \ldots, \bar{W}_M).
  \label{eq:app-symmetry}
\end{equation}
Two configurations of $(\alpha, \beta)$ with the same product $\beta\alpha^{M}$ therefore
compute the same network function, with the same input--output Jacobian and effective rank
(Figure~\ref{fig:rank-symmetry}). In Pre-Norm the normalization acts on the branch input,
before the first branch matrix, and in Post-Norm it acts on the stream after the residual
addition; in both cases the branch itself is a product of matrices and activations, so the
argument applies to either placement.

The equivalence persists throughout training if the optimizer hyperparameters are rescaled
accordingly for each matrix. To see why, rescale $\alpha \gets \lambda\alpha$ and
$\beta \gets \beta/\lambda^{M}$ for some $\lambda > 0$, which preserves $\beta\alpha^{M}$.
The forward pass is unchanged, but each branch matrix is $\lambda$ times larger and its
gradient is $\lambda$ times smaller: the gradient with respect to one matrix carries the
branch scale $\beta/\lambda^{M}$ and the other $M{-}1$ matrices, which contribute
$\lambda^{M-1}$. For the rescaled network to follow the same trajectory, its matrices must
remain $\lambda$ times larger after every update, so the learning rate must be rescaled in
accordance with how the gradient changed: $\eta \gets \lambda^{2}\eta$ for SGD, whose
update is proportional to the gradient, and $\eta \gets \lambda\eta$ for Adam, whose update
is invariant to the gradient scale. Similar symmetries were observed for MLP layers in
\citet{yang2021tensoriv,yang2023tensorivb} and for the Pre-Norm residual stream in
\citet{blake2024umup}. The rescaling must extend to other optimizer
hyperparameters, including $\epsilon$ in Adam, weight decay, and gradient clipping
\citep{yang2023tensorivb}. This equivalence holds exactly only in infinite precision and
infinite dynamic range.

\section{Normalization Placement: Additional Results and Derivations}
\label{app:norm}

\subsection{Block Jacobians and the normalization projection}
\label{app:norm-deriv}

We consider a residual block that may contain a normalization operation in each of three
locations \citep{kim2025peri}: the branch input ($P$), the branch output ($B$), and the
residual stream after the addition ($R$). With branch $f$, the block maps
\begin{equation}
  h_\ell = R\big(h_{\ell-1} + \beta\,B(f(P(h_{\ell-1})))\big).
\end{equation}
By the chain rule its Jacobian factorizes as
\begin{equation}
  \frac{\partial h_\ell}{\partial h_{\ell-1}} = J_R\,\big(I + \beta\,J_B J_f J_P\big),
  \label{eq:block-jac}
\end{equation}
where each $J$ is the Jacobian of its operation at the relevant point. The placement
determines where a normalization Jacobian lands: a branch operation ($P$ or $B$) sits inside
the $\beta$-scaled branch term, multiplied by the small ratio, whereas the post-residual
operation $R$ multiplies the entire sum, the skip included. This is the structural reason
placement matters --- only $R$ applies its Jacobian to the identity path.

\paragraph{The RMSNorm Jacobian is a scaled projection.}
With its gain at the standard initialization of one, RMSNorm is $N(x) = \sqrt d\,x/\|x\|$,
and
\begin{equation}
  J_N = \frac{\partial N}{\partial x}
      = \frac{\sqrt d}{\|x\|}\big(I - \hat x\hat x^\top\big),
  \qquad \hat x = \frac{x}{\|x\|}.
  \label{eq:rmsnorm-jac}
\end{equation}
The factor $\sqrt d/\|x\|$ is the inverse stream scale (the root-mean-square of $x$ is
$\|x\|/\sqrt d$); the projection $I - \hat x\hat x^\top$ removes the current stream direction
$\hat x$.

\paragraph{The detached rescaling separates the scale from the projection.}
To separate the effect of the projection term from the effect of the scalar rescaling, we
define a \emph{detached rescaling} that divides by the stream scale exactly as RMSNorm does
but stops the gradient through that scale: it computes $x/\operatorname{sg}(\mathrm{rms}(x))$,
where $\operatorname{sg}$ is the stop-gradient. The forward pass is identical to RMSNorm, but
the scale is now a constant, so the Jacobian is
\begin{equation}
  J_{\mathrm{detach}} = \frac{\sqrt d}{\|x\|}\,I,
  \label{eq:detach-jac}
\end{equation}
the same inverse scale with no projection. It rescales the stream but removes no direction,
which is what makes it the control that isolates the rescaling from the projection
(Appendix~\ref{app:norm-control}).

\paragraph{The four post-residual operations.}
In Table~\ref{tab:norm-jac} we summarize the four operations we use in the post-residual
placement. They let us vary the two factors of a normalization Jacobian independently: the
scalar rescaling $\sqrt d/\|x\|$, which normalizes the stream to unit scale, and the projection
$I - (\cdot)(\cdot)^\top$, which removes a direction. RMSNorm does both; the detached rescaling
keeps only the rescaling; mean subtraction keeps only the projection, of a fixed direction and
without rescaling; LayerNorm adds mean removal to RMSNorm. Comparing them isolates the
contribution of each factor (Appendix~\ref{app:norm-control}).

\begin{table}[ht]
  \centering
  \caption{Jacobians of the post-residual operations at gain one. $\hat x = x/\|x\|$;
  $\hat x_c$ is the unit centered vector; $\mathbf 1$ is the all-ones direction.}
  \label{tab:norm-jac}
  {\small
  \begin{tabular}{lccc}
    \toprule
    operation & Jacobian & rescales? & removes \\
    \midrule
    RMSNorm & $\frac{\sqrt d}{\|x\|}\big(I - \hat x\hat x^\top\big)$ & yes & radial $\hat x$ \\[2pt]
    detached rescale & $\frac{\sqrt d}{\|x\|}\,I$ & yes & nothing \\[2pt]
    mean subtraction & $I - \frac1d\mathbf 1\mathbf 1^\top$ & no & fixed $\mathbf 1$ \\[2pt]
    LayerNorm & $\frac{\sqrt d}{\|x-\mu\|}\big(I - \hat x_c\hat x_c^\top\big)\big(I - \frac1d\mathbf 1\mathbf 1^\top\big)$ & yes & radial $+\,\mathbf 1$ \\
    \bottomrule
  \end{tabular}}
\end{table}

\subsection{The projection is not the primary cause of rank loss}
\label{app:norm-control}

In Section~\ref{sec:norm-placement} we observed that the Jacobians of the normalization
operations contain projection terms that remove a direction from the gradient: mean
subtraction removes the all-ones direction $\mathbf 1$, and rescaling removes the current
stream direction $\hat x = h/\|h\|$,
\[
  \frac{\partial}{\partial h}\bigl(h - \mu(h)\mathbf{1}\bigr) = I - \tfrac{1}{d}\mathbf{1}\mathbf{1}^{\top},
  \qquad
  \frac{\partial}{\partial h}\,\frac{h}{\mathrm{RMS}(h)} = \frac{\sqrt{d}}{\|h\|}\bigl(I - \hat{x}\hat{x}^{\top}\bigr)
\]
(Eq.~\ref{eq:norm-proj}). Composed over $L$ layers, these projections could in principle
compound into a low-rank Jacobian, which makes the projection terms a natural candidate for
explaining the rank collapse \citep{emadi2026exact}. Here we expand on the main-text argument that the projection terms instead have a
negligible effect on the rank, and that the branch-to-skip ratio is the primary factor
governing the rate of rank loss: we show that the projection is neither necessary nor
sufficient for the collapse.

\paragraph{The projection is not necessary for rank collapse.}
We use the detached rescaling operation to ablate the projection term (Eq.~\ref{eq:detach-jac}),
which lets us test whether the projection is necessary for the rank collapse to occur. Even with
the projection removed, the detached rescaling collapses the rank just like ordinary RMSNorm. Across $\alpha\in\{0.5,1,2\}$,
$L\in\{48,\dots,384\}$, and $\beta\in\{1,\,1/\sqrt L\}$, with and without the Pre-RMSNorm, the
detached and ordinary ranks differ by at most $0.4\%$ of the full rank (largest $0.0039$, at
$\alpha{=}0.5$ and $\beta{=}1/\sqrt L$, where the rank is near full).
Figure~\ref{fig:norm-mechanism-dial}b overlays the detached and ordinary curves across
depth. The projection is therefore not necessary for the collapse.

\paragraph{The projection is not sufficient for rank collapse.}
Now, to simulate what a projection term without the rescaling would do, we form the product
$\prod_{\ell}(I - x_\ell x_\ell^\top)$ of rank-one projections, where $x_\ell$ follows an
independent random walk on the sphere with step angle $\epsilon$, so each layer removes a
different, uncorrelated direction --- the worst case for compounding. The effective rank does
not collapse: it stays at $0.996$ of the width across step sizes $\epsilon\in[0.01,0.6]$ and
depths $L\in[48,384]$, falling only to $0.98$ at the largest step and depth
(Table~\ref{tab:surrogate}). Because each removed direction is an independent random vector, it
generically does not lie in the range of the accumulated product, so it removes no new dimension
and only attenuates the smallest singular values: the product keeps rank $d-1$ at every depth
instead of losing a dimension per layer. Mean subtraction is the opposite extreme, removing the
same fixed direction at every layer, which can cost at most one dimension. Direction removal
collapses the rank at neither extreme: not when each layer removes an independent direction, and
not when every layer removes the same one.

\begin{table}[ht]
  \centering
  \caption{Effective rank over width of a product of $L$ rank-one projections along an
  independent random walk of step angle $\epsilon$ ($d{=}256$, mean over $2$ seeds). The rank
  does not collapse, so removing a direction per layer is not by itself the cause.}
  \label{tab:surrogate}
  \begin{tabular}{lcccc}
    \toprule
    $\epsilon$ & $L{=}48$ & $L{=}96$ & $L{=}192$ & $L{=}384$ \\
    \midrule
    $0.01$ & $0.996$ & $0.996$ & $0.996$ & $0.996$ \\
    $0.15$ & $0.996$ & $0.996$ & $0.996$ & $0.996$ \\
    $0.30$ & $0.996$ & $0.996$ & $0.995$ & $0.995$ \\
    $0.60$ & $0.994$ & $0.992$ & $0.988$ & $0.980$ \\
    \bottomrule
  \end{tabular}
\end{table}

The collapse is therefore governed by the rescaling, not the projection: keeping the
rescaling without the projection still shows rank collapse, whereas applying the projection
without the rescaling does not induce the rank collapse.

\begin{figure}[ht]
  \centering
  \includegraphics[width=\textwidth]{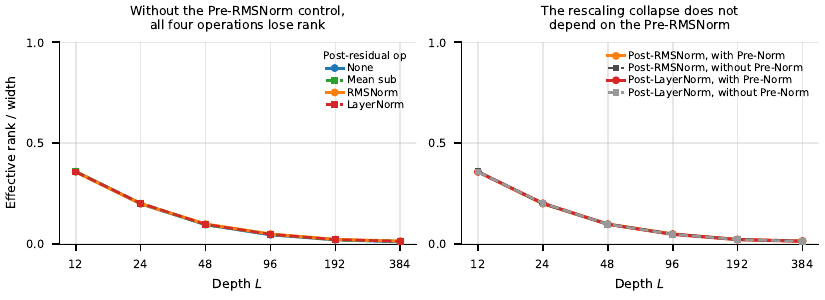}
  \caption{\textbf{Without the Pre-RMSNorm control, the unnormalized configurations collapse
  and mask the operation comparison.} Effective rank of the input--output Jacobian across
  depth at $\alpha{=}\beta{=}1$, as in Figure~\ref{fig:norm-mechanism-dial}b but with no
  Pre-RMSNorm. \textbf{(a)}~With no operation or mean subtraction alone, there is no
  normalization on the branch: the branch-to-skip ratio stays constant and the rank collapses
  at nearly the rate of the rescaling operations, so the four operations are indistinguishable.
  \textbf{(b)}~The rescaling collapse does not depend on the control: the post-RMSNorm and
  post-LayerNorm curves with and without the Pre-RMSNorm coincide.}
  \label{fig:norm-nopre-control}
\end{figure}

\paragraph{Isolating the post-residual operation.}
Figure~\ref{fig:norm-mechanism-dial}b adds a Pre-RMSNorm to every configuration so that the
four post-residual operations are compared on a normalized branch input, varying only the
operation. Without a Pre-RMSNorm to control the magnitude, the configurations with no
post-residual operation or with mean subtraction have no normalization on the branch at all:
their branch-to-skip ratio stays constant, so they collapse at nearly the rate of the rescaling
operations and the four become indistinguishable (Figure~\ref{fig:norm-nopre-control}a). To
show that this Pre-RMSNorm does not confound our results, Figure~\ref{fig:norm-nopre-control}b
shows that with or without it the post-RMSNorm and post-LayerNorm curves coincide.

\subsection{The depth law: the ratio across depth, contour extraction, and width independence}
\label{app:norm-law}

\paragraph{The branch-to-skip ratio.}
The branch-to-skip ratio (Eq.~\ref{eq:branch-skip}) is the Frobenius norm of the branch
term in the block Jacobian (Eq.~\ref{eq:block-jac}) relative to that of the identity skip. At initialization the branch term
$\beta\,J_B J_f J_P$ collects three factors: the explicit branch scale $\beta$; a factor
$\alpha^M$ from the $M$ branch weight matrices, each initialized at scale $\alpha$, so the
branch Jacobian $J_f$ scales as $\alpha^M$; and the inverse stream scale
$\sqrt d/\|h_\ell\|$ contributed by the branch normalization (Eq.~\ref{eq:rmsnorm-jac}),
from the branch-input normalization in Pre-Norm or, for a homogeneous branch, equivalently
from the branch-output normalization. Their product is
\begin{equation}
  \tilde r_\ell = \frac{\beta\alpha^M\sqrt d}{\|h_\ell\|}
         = \frac{\beta\alpha^M}{\|h_\ell\|/\sqrt d},
  \label{eq:ratio}
\end{equation}
the nominal ratio $\beta\alpha^M$ divided by the stream scale
$\mathrm{rms}(h_\ell) = \|h_\ell\|/\sqrt d$. As in the main text, we track the nominal form
$\tilde r_\ell = r_\ell/\sqrt{p}$, dropping the constant activation-derivative factor
$\sqrt{p}$ (Section~\ref{sec:skip}), which is shared by every block and both placements.
The nominal ratio $\beta\alpha^M$ is $\tilde r_\ell$ at unit stream scale
($\|h_\ell\| = \sqrt d$); the per-layer ratio scales inversely with how far the stream
has grown. All of the depth dependence enters through $\|h_\ell\|$: a post-residual
rescaling resets it to $\sqrt d$ at every block, whereas the branch normalizations leave the
stream to grow with depth.

\begin{figure}[ht]
  \centering
  \includegraphics[width=\textwidth]{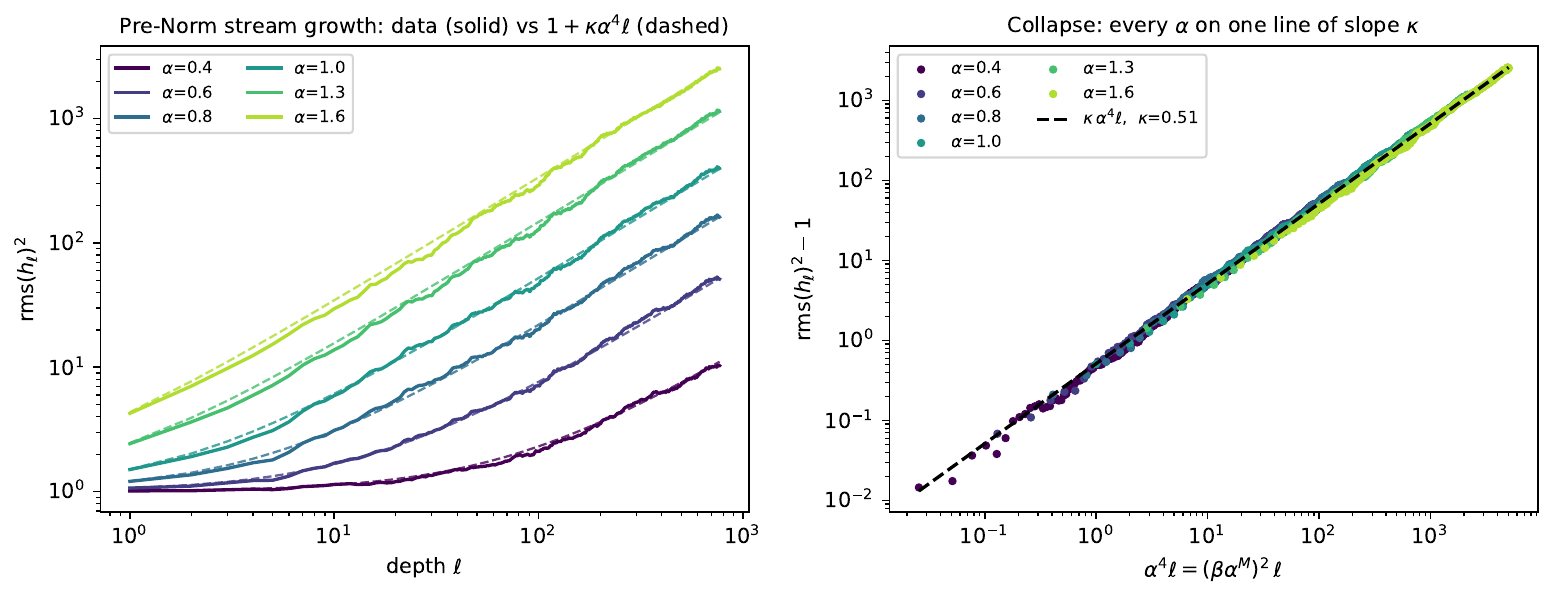}
  \caption{\textbf{At initialization, $\mathrm{rms}(h_\ell)^2$ grows like
  $1+\kappa(\beta\alpha^M)^2\ell$ with $\kappa = \mathbb{E}[\sigma^2] = \tfrac{1}{2}$ for
  ReLU; the measured $\kappa\approx0.51$ agrees at every initialization scale.} \textbf{(a)}~$\mathrm{rms}(h_\ell)^2$ versus depth for
  a Pre-Norm two-matrix ReLU branch (both matrices $d\times d$, $d{=}256$; $\beta{=}1$),
  shown as the measured stream
  (solid) against the fit (dashed); the stream grows faster at larger $\alpha$.
  \textbf{(b)}~Plotted against $(\beta\alpha^M)^2\ell$, $\mathrm{rms}(h_\ell)^2-1$ collapses
  onto a single line of slope $\kappa\approx0.51$ for all $\alpha$, confirming
  $\|h_\ell\|^2 = d(1+\kappa(\beta\alpha^M)^2\ell)$.}
  \label{fig:norm-stream}
\end{figure}

\paragraph{Residual stream growth.}
When the skip connection is an identity (which occurs whenever there is no post-residual
normalization), the residual stream is a running sum
$h_\ell = h_{\ell-1} + \beta u_\ell$ of branch outputs $u_\ell$. The branch input is
normalized, so each $u_\ell$ has roughly fixed magnitude and, at initialization, is nearly
orthogonal to the accumulated stream (two generic vectors in $d$ dimensions have normalized
inner product $O(1/\sqrt d)$). Expanding,
\[
  \|h_\ell\|^2 = \|h_{\ell-1}\|^2 + 2\beta\langle h_{\ell-1}, u_\ell\rangle
              + \beta^2\|u_\ell\|^2,
\]
the cross term vanishes exactly in expectation, since $W_{\mathrm{down}}$ is zero-mean and
independent of the rest of the branch, and the increment $\beta^2\|u_\ell\|^2$ is constant
in expectation. The squared length therefore grows linearly (the magnitude recurrence in
Appendix~\ref{app:skip}): $\|h_\ell\|^2 = d\,(1+\kappa(\beta\alpha^M)^2\ell)$. The
coefficient is $\kappa = \mathbb{E}[\sigma(z)^2]$, the second moment of the activation
values on a unit-variance input, equal to $\tfrac{1}{2}$ exactly for ReLU. Block $\ell$'s branch sees the stream $h_{\ell-1}$,
so the branch-to-skip ratio is
\begin{equation}
  \tilde r_\ell = \frac{\beta\alpha^{M}}{\sqrt{1+\kappa(\beta\alpha^{M})^{2}(\ell-1)}},
  \label{eq:ratio-decay}
\end{equation}
which starts near $\beta\alpha^M$
and decays as $1/\sqrt\ell$ as the stream grows, whereas a post-residual rescaling resets
$\|h_\ell\| = \sqrt d$ every block and holds the ratio constant at $\beta\alpha^M$. The measured value, $\kappa\approx0.51$ for a ReLU branch, matches the
derivation: $\mathrm{rms}(h_\ell)^2-1$ collapses onto a single line in
$(\beta\alpha^M)^2\ell$ (Figure~\ref{fig:norm-stream}). $\kappa$ is a moment of the
activation values $\sigma(z)$, which determine the branch outputs accumulating on the
stream. The branch Jacobians are controlled by the derivatives moment
$p = \mathbb{E}[\sigma'(z)^2]$ (Section~\ref{sec:skip}). For ReLU both moments equal
$\tfrac{1}{2}$, by the Gaussian identity $\mathbb{E}[z^2\mathbf{1}(z>0)] = P(z>0)$.
At $L{=}384$, $\alpha{=}\beta{=}1$, the
stream reaches $\mathrm{rms}(h_L)\approx 14.5$ for the Pre-Norm input and $\approx 19.7$ for a
unit-increment branch-output normalization, versus $1.00$ for the post-residual placement.

\paragraph{Networks with post-residual rescaling behave like networks with no normalization.}
For a degree-1 positively-homogeneous activation such as ReLU, $f(ch)=cf(h)$ for $c>0$, so
the branch Jacobian is scale-invariant: $J_f(ch)=J_f(h)$. This has two consequences. First,
a network with no normalization keeps the same stream \emph{direction} as Post-Norm: by
induction, if $h^{\mathrm{none}}_{\ell-1} = s_{\ell-1}\,\hat h^{\mathrm{post}}_{\ell-1}$ for a
positive scalar $s_{\ell-1}$, homogeneity makes the branch outputs match up to $s_{\ell-1}$,
so $h^{\mathrm{none}}_\ell$ is again a scalar multiple of the Post-Norm sum. The two
placements feed their branches the same direction, hence the same $J_f$. Second, the detached
rescaling is the no-norm Jacobian up to a per-layer scalar: its block Jacobian is
$\frac{\sqrt d}{\|\cdot\|}(I + \beta J_f)$ (Eq.~\ref{eq:detach-jac}), a scalar times the
no-norm block Jacobian $I + \beta J_f$. The scalars factor out of the product and do not
affect the rank, so the detached rescaling and no normalization have identical Jacobian
effective rank.

Ordinary Post-Norm adds the projection $I - \hat x\hat x^\top$ to each block
(Eq.~\ref{eq:rmsnorm-jac}), the only difference from the detached and no-norm cases. Removing
that projection (the detached rescaling, Appendix~\ref{app:norm-control}) changes the
effective rank by at most $0.4\%$ of the width, so the no-norm, detached, and Post-Norm ranks
all coincide to within that margin, and the branch-to-skip ratio is constant $\beta\alpha^M$
for all three. The degree-1 homogeneity is essential: a degree-2 activation would make $J_f$
scale as $\|h\|$, so the no-norm ratio would grow with depth instead of staying constant.

\paragraph{The $\sqrt{L}$ depth factor from path counting.}
The path decomposition in Section~\ref{sec:skip} (Eq.~\ref{eq:path-decomp}) writes the
input--output Jacobian as a sum over path orders, where the order-$k$ paths pass through the
branch at $k$ blocks and through the skip everywhere else. For Pre-Norm the decomposition
applies directly; for Post-Norm we first drop the post-residual norm Jacobians, whose scalar
factors do not change the effective rank and whose projections empirically contribute at most
$0.4\%$ of the width (Appendix~\ref{app:norm-control}). Relative to the Frobenius norm
$\sqrt{d}$ of the identity skip, a path that visits the branch at blocks
$\ell_1 < \cdots < \ell_k$ has Frobenius norm $r_{\ell_1}\cdots\, r_{\ell_k}$, the product of
the branch-to-skip ratios at the visited blocks, up to an order-one constant per factor that
is the same for every $\alpha$, $\beta$, and $L$.

At initialization, the branch Jacobians at different blocks are independent, so distinct
paths point in nearly uncorrelated directions and their squared norms add. The order-$k$
path sum then has relative Frobenius norm
\begin{equation}
  \Big(\sum_{\ell_1<\cdots<\ell_k} r_{\ell_1}^{2}\cdots\, r_{\ell_k}^{2}\Big)^{1/2}
  \;\approx\; \frac{\big(\sum_\ell \tilde r_\ell^2\big)^{k/2}}{\sqrt{k!}},
  \label{eq:orderk-norm}
\end{equation}
where the approximation holds when the individual $\tilde r_\ell^2$ are small. For
Post-Norm, $\tilde r_\ell = \beta\alpha^{M}$ at every block, so
$\sum_\ell \tilde r_\ell^2 = (\beta\alpha^{M})^2 L$ and
the order-$k$ sum has norm $(\beta\alpha^{M}\sqrt{L})^{k}/\sqrt{k!}$. Each branch visit
multiplies a path's norm by $\beta\alpha^{M}$ and multiplies the number of available paths by
roughly $L$; a sum of $N$ randomly oriented terms of equal norm grows as $\sqrt{N}$ times
that norm, so each visit contributes one factor of $\beta\alpha^{M}\sqrt{L}$. For Pre-Norm,
the ratios decay following Eq.~\ref{eq:ratio-decay}, so
\begin{equation}
  \sum_{\ell=1}^{L} \tilde r_\ell^{2}
  \;\approx\; \int_{0}^{L} \frac{(\beta\alpha^{M})^{2}}{1+\kappa(\beta\alpha^{M})^{2}\ell}\,d\ell
  \;=\; \frac{1}{\kappa}\,\log\!\big(1+\kappa(\beta\alpha^{M})^{2}L\big),
  \label{eq:pre-sumr2}
\end{equation}
again a function of $\beta\alpha^{M}\sqrt{L}$ alone, since
$(\beta\alpha^{M})^{2}L = (\beta\alpha^{M}\sqrt{L})^{2}$. In the Frobenius ratio $r_\ell = \sqrt{p}\,\tilde r_\ell$, each
order carries an additional factor $(\sqrt{p})^{k}$ and Eq.~\ref{eq:pre-sumr2} a prefactor
$p/\kappa$; these shift the collapse onset by the constant $1/\sqrt{p}$, identically for
both placements.

Under three assumptions, which we state below along with measurements that validate them,
every order of the expansion depends on $(\alpha,\beta,L)$ only through
$\beta\alpha^{M}\sqrt{L}$; since the effective rank is scale-invariant, it is then a function
of this single variable for each placement, and the iso-rank contours scale as
$\beta\alpha^{M} \propto 1/\sqrt{L}$. First, the count $\binom{L}{k}\approx L^{k}/k!$ requires
$k \ll L$; near the transition at $\beta\alpha^{M}\sqrt{L} \sim 1$ the contributing orders are
small, so this condition is mild. Second, distinct paths are treated as uncorrelated. Each
branch Jacobian ends in a zero-mean output matrix, which suppresses correlations between
paths, but the activation masks depend on the shared stream, so weak correlations remain; the
measured exponents test the net effect, since coherently adding paths would grow the
order-$k$ sum as $(\beta\alpha^{M}L)^{k}/k!$ and give iso-rank contours scaling as $1/L$,
whereas the measured contour slope is $-0.50$ (Figure~\ref{fig:norm-law}c) and the order-one
sum grows with log--log slope $0.50$ (Figure~\ref{fig:norm-law}b), matching the uncorrelated
prediction. Third, the Frobenius norms of the path sums do not by themselves determine the
rank: we additionally assume that the path sums, normalized by their Frobenius norms, have
distributions that do not otherwise depend on $\alpha$, $\beta$, or $L$ (their spectra and
their alignments with one another and with the identity). For homogeneous activations like
ReLU, $\alpha$ and $\beta$ enter each path exactly as the overall factor
$(\beta\alpha^{M})^{k}$, so this assumption concerns the depth dependence alone for
Post-Norm, and the profile of the decaying ratios for Pre-Norm; the collapse of the measured
rank curves onto a single function of $\beta\alpha^{M}\sqrt{L}$ (Figure~\ref{fig:norm-law}d)
empirically validates that no further dependence on $\alpha$, $\beta$, or $L$ remains. We do
not attempt a closed form for the shape of the rank curve: a simple independent-increment
model of the spectrum does not match the measurements, with the log-singular-value spread
growing faster with depth than the leading order predicts.

\begin{figure}[p]
  \centering
  \includegraphics[width=\textwidth,height=0.72\textheight,keepaspectratio]{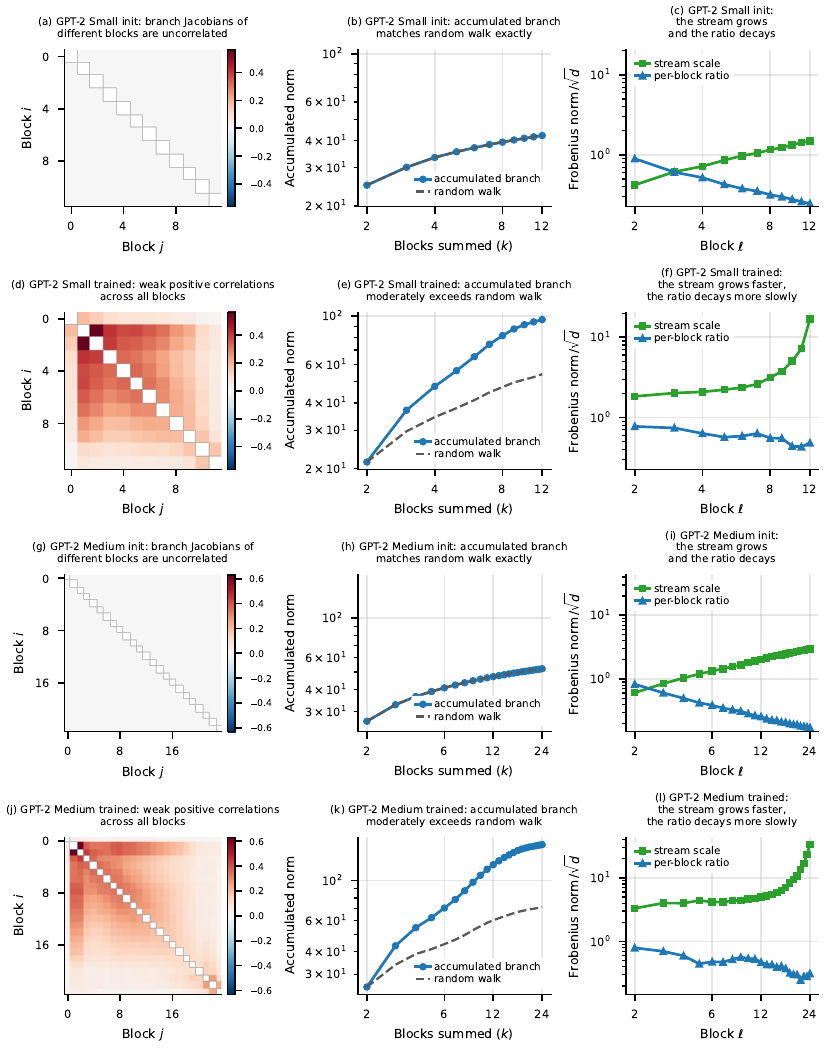}
  \caption{\textbf{The assumptions behind the $\sqrt{L}$ depth factor hold exactly in
  GPT-2 at initialization; after training, the path decorrelation is mildly violated and
  the stream grows faster.} The path-counting argument assumes that the FFN branch
  Jacobians of different blocks are uncorrelated, so that the path sums of
  Eq.~\ref{eq:orderk-norm} accumulate as random walks, and the Pre-Norm argument depends
  on the residual stream growing while the branch-to-skip ratio decays
  (Eq.~\ref{eq:ratio-decay}). All panels: FFN branch Jacobians $J_\ell$ computed
  analytically along the residual stream of a fixed prompt, at random initialization and
  with the pretrained weights, for GPT-2 Small ($d{=}768$, $L{=}12$, rows one and two)
  and GPT-2 Medium ($d{=}1024$, $L{=}24$, rows three and four). Block~1 is omitted from
  the curve panels: its small incoming stream inflates $\|J_1\|$ through the
  normalization Jacobian.
  \textbf{(a,d,g,j)}~Pairwise cosine similarities $\langle J_i,
  J_j\rangle/\|J_i\|\|J_j\|$ between blocks; the random-direction baseline is
  ${\approx}1/d$.
  \textbf{(b,e,h,k)}~The accumulated branch norm $\|\sum_{\ell\le k} J_\ell\|_F$ as
  blocks are summed, against the prediction for uncorrelated blocks,
  $\smash{(\sum_{\ell\le k}\|J_\ell\|^2)^{1/2}}$, computed from the measured per-block
  norms (dashed): equal at initialization, moderately exceeded after training.
  \textbf{(c,f,i,l)}~The stream scale $\mathrm{rms}(h_\ell)$ and the per-block ratio
  $r_\ell = \|J_\ell\|_F/\sqrt{d}$ (the Frobenius ratio in Eq.~\ref{eq:branch-skip}): the
  stream grows and the ratio decays in both cases,
  with the trained growth concentrated in the final blocks.}
  \label{fig:norm-gpt2-assumptions}
\end{figure}

\paragraph{The assumptions on a real transformer, before and after training.}
We test these assumptions on GPT-2 Small and GPT-2 Medium, computing the FFN branch
Jacobians analytically along the residual stream of a fixed prompt, both at random
initialization and with the pretrained weights (Figure~\ref{fig:norm-gpt2-assumptions}).
At initialization, every assumption holds exactly: the pairwise cosine similarities
between blocks sit at the random-direction baseline of ${\approx}1/d$, and the accumulated
branch norm equals the uncorrelated prediction ($1.00\times$ at $k{=}L$ for both models).
Training induces weak, uniformly positive correlations between blocks (off-diagonal mean
$0.18$ for GPT-2 Small and $0.15$ for GPT-2 Medium). The accumulated branch reaches
$1.78\times$ and $2.23\times$ the uncorrelated prediction, against $3.26\times$ and
$4.59\times$ for fully coherent addition: roughly half of the fully coherent value in
both models. The
stream growth and the ratio decay persist in direction after training, but the trained
stream grows faster, with the growth concentrated in the final blocks.
\citet{sun2025curseofdepth} measure the same concentration during pretraining of a
12-layer model, and bound the Pre-LN stream variance between $\Theta(L)$ and
$\Theta(\exp L)$ under Gaussian independence assumptions; our
initialization-versus-trained comparison locates the two ends of this range empirically,
with the initialized stream at the linear lower bound, which under their assumptions
corresponds to an uncorrelated branch and stream. A faster growing stream suppresses the
branch-to-skip ratio further, pushing the deep layers toward the skip-dominated regime
rather than toward rank collapse. Two caveats: GPT-2 uses GELU and its stream includes
the attention sublayers, so these are measurements of the assumptions rather than of our
exact ReLU setting; and the trained per-block profiles vary with the probe token
position, whereas the correlation and accumulation measurements do not.

We measure the depth law in Figure~\ref{fig:norm-law} as follows. The branch-scale symmetry
(Figure~\ref{fig:rank-symmetry}) lets us vary a single quantity: we fix $\beta = 1$ and sweep
the initialization scale $\alpha$, so that $\beta\alpha^M = \alpha^M$. At each depth $L$ and
each rank level $r^\star$, we extract the \emph{iso-rank contour} --- the value of
$\beta\alpha^M$ at which the effective rank equals $r^\star$ --- by interpolating
$\log\beta\alpha^M$ against the measured rank. Fitting $\log\beta\alpha^M$ against $\log L$
then gives the contour's power-law exponent.

The exponent is close to $-\tfrac12$ across rank levels and both placements
(Table~\ref{tab:law-slopes}), the $1/\sqrt L$ scaling reported in the main text. The Post-Norm
contours are the cleanest, with slopes between $-0.49$ and $-0.52$; the Pre-Norm contours are
noisier because the rank plateaus near its floor, which flattens the rank--$\alpha$ curve and
makes the crossing harder to locate, but they remain near $-0.5$.

A wider branch does not change the exponent. Repeating the fit with a $4\times$ width expansion
(a hidden dimension four times the model dimension, both branches having $M=2$ matrices) leaves
the slopes near $-0.5$ (Table~\ref{tab:law-slopes}, $4\times$). The width raises the rank floor
at every depth --- the per-block conditioning effect in Section~\ref{sec:width} --- but the
depth scaling is unchanged, so the $\sqrt L$ factor is a property of depth rather than of the
per-block branch.

\begin{table}[ht]
  \centering
  \caption{Fitted exponent $s$ of the iso-rank contour $\beta\alpha^M \propto L^{s}$, for both
  placements, three rank levels $r^\star$, and two width expansions ($1\times$, the
  two-matrix branch used throughout; $4\times$, a four-times-wider hidden dimension). All are
  near $s=-\tfrac12$.}
  \label{tab:law-slopes}
  \begin{tabular}{llccc}
    \toprule
    placement & width & $r^\star{=}0.4$ & $r^\star{=}0.5$ & $r^\star{=}0.6$ \\
    \midrule
    Pre-Norm  & $1\times$ & $-0.46$ & $-0.50$ & $-0.52$ \\
    Pre-Norm  & $4\times$ & $-0.57$ & $-0.47$ & $-0.48$ \\
    Post-Norm & $1\times$ & $-0.51$ & $-0.50$ & $-0.49$ \\
    Post-Norm & $4\times$ & $-0.52$ & $-0.50$ & $-0.50$ \\
    \bottomrule
  \end{tabular}
\end{table}

\subsection{Trainability: the learning-rate basin}
\label{app:norm-lr}

The trainability grid in Figure~\ref{fig:norm-trainability} reports the best CIFAR-10 test
accuracy over an Adam learning-rate sweep ($\eta$ from $3\times10^{-5}$ to $10^{-2}$) at each
initialization scale $\alpha$ and depth $L$. We also define the maximum stable learning
rate as the largest $\eta$ at which the network still trains, which we take to be a test
accuracy of at least $0.4$ against a chance level of $0.1$. Table~\ref{tab:lr-basin} reports the
maximum stable learning rate across depth.

Pre-Norm is able to train with all learning rate values in our sweep (up to $\eta = 10^{-2}$) at
every depth and every $\alpha$, consistent with its rank floor. The stable learning rate range
for Post-Norm instead narrows with depth, and faster at larger $\alpha$
(Table~\ref{tab:lr-basin}).

The narrowing basin is an early signal of the trainability failure. At $\alpha = 0.5$ the
maximum stable learning rate falls by more than two orders of magnitude across depth (from
$10^{-2}$ to $3\times10^{-5}$), yet the best accuracy over the sweep stays at $0.50$ or above
at every depth: the network still trains at a small enough learning rate. At $\alpha = 1$ the
maximum stable learning rate falls thirtyfold by depth 96, and the best accuracy first drops
at depth 192 ($0.18$), reaching chance at depth 384. The learning-rate basin therefore narrows at a smaller depth than the depth where a loss of
accuracy on CIFAR first appears.

\begin{table}[ht]
  \centering
  \caption{$\log_{10}$ of the maximum stable Adam learning rate (the largest $\eta$ with
  CIFAR-10 test accuracy $\ge 0.4$; chance is $0.1$) versus depth, at $\beta = 1$, from the sweep behind
  Figure~\ref{fig:norm-trainability}. Pre-Norm stays at the top of the sweep at every depth;
  Post-Norm's stable range narrows with depth, faster at larger $\alpha$. A dash marks a
  configuration that does not reach $0.4$ at any learning rate in the sweep
  ($3\times10^{-5}$ to $10^{-2}$).}
  \label{tab:lr-basin}
  \begin{tabular}{llcccccc}
    \toprule
    placement & $\alpha$ & $L{=}12$ & $24$ & $48$ & $96$ & $192$ & $384$ \\
    \midrule
    Pre-Norm  & all & $-2$ & $-2$ & $-2$ & $-2$ & $-2$ & $-2$ \\
    Post-Norm & $0.5$ & $-2$ & $-2$ & $-2.5$ & $-3$ & $-3.5$ & $-4.5$ \\
    Post-Norm & $1$ & $-2.5$ & $-3$ & $-3.5$ & $-4$ & --- & --- \\
    Post-Norm & $2$ & $-2.5$ & $-3.5$ & $-4$ & --- & --- & --- \\
    Post-Norm & $4$ & $-2$ & $-3$ & $-4$ & --- & --- & --- \\
    \bottomrule
  \end{tabular}
\end{table}

\subsection{Trainability of Output-Norm}
\label{app:norm-slots}

\begin{figure}[ht]
  \centering
  \includegraphics[width=0.62\textwidth]{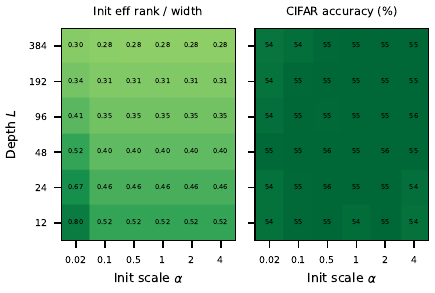}
  \caption{\textbf{Output-Norm trains across the grid, like Pre-Norm.}
  Initialization effective rank over width (left) and best CIFAR-10 test accuracy over an Adam
  learning-rate sweep (right) for a normalization at the branch output ($\beta{=}1$), over the
  same grid as Figure~\ref{fig:norm-trainability}. The rank stays above collapse and the
  accuracy stays near $55\%$ at every initialization scale $\alpha$ and depth $L$, matching
  Pre-Norm (Figure~\ref{fig:norm-trainability}a,b).}
  \label{fig:norm-bslot}
\end{figure}

Figure~\ref{fig:norm-mechanism-dial}a shows that both branch placements --- a normalization at
the branch input (Pre-Norm) or the branch output --- keep a rank floor, while only the
post-residual placement collapses. Output-Norm sits at a marginally lower floor
than Pre-Norm, visible in Figure~\ref{fig:norm-mechanism-dial}a where its curve runs just below
Pre-Norm's: a branch-output normalization fixes every branch increment to unit scale, whereas
Pre-Norm's branch-input normalization feeds the branch a unit input and the ReLU branch
contracts it slightly, to a root-mean-square of about $0.74$. Output-Norm
therefore adds a slightly larger increment at each block and runs at a marginally larger
branch-to-skip ratio, which lowers its floor. Output-Norm still maintains rank above the threshold where it would harm CIFAR
accuracy: across the same grid of initialization scale and depth, its CIFAR-10 accuracy stays
near $55\%$ everywhere, similar to Pre-Norm (Figure~\ref{fig:norm-bslot}), unlike Post-Norm,
which fails at large $\alpha$ and depth (Figure~\ref{fig:norm-trainability}).
A branch-output normalization can still destabilize the residual-stream magnitude at large
depth \citep{kim2025peri}, but that is a magnitude effect, separate from the rank-driven
failure studied here.

\section{Branch Coherence: Derivations and Measurements}
\label{app:coherence}

\paragraph{The cross-layer correlation equals $c_\ell$ (spiked regime).}
We derive the main-text identity $\cos(m_\ell, m_{\ell'}) = c_\ell$ for a deep
network at initialization, in the regime where the mean spike dominates the
residual stream. The branch mean at layer $\ell$ is
$m_\ell = \mathbb{E}_x[\sigma(W_\ell z)]$, the average of the branch output over
inputs, with $z$ the Pre-Normed branch input and $W_\ell$ isotropic. Once the
spike dominates, the representations of all inputs have collapsed onto its
direction, which is the all-ones direction $\mathbf 1$: an uncentered activation
gives every coordinate of the branch output the same positive mean, so each
branch mean carries a $\mathbf 1$-component, and these accumulate until
$\cos(\mu_\ell, \mathbf 1)\to 1$ (verified below). After the Pre-Norm every input
then has $z \approx \mathbf 1$, independent of $x$, so the pre-activation of
coordinate $i$ is
\[
  (W_\ell z)_i \;\approx\; (W_\ell \mathbf 1)_i \;=:\; g_i^{(\ell)} .
\]
The rows of $W_\ell$ are independent and the Pre-Norm fixes the scale, so the
$g_i^{(\ell)}$ are i.i.d.\ standard Gaussian and independent across layers. The
average over inputs has collapsed --- the pre-activation no longer depends on $x$
-- so each branch mean is a deterministic function of that layer's weights,
$m_\ell[i] \approx \sigma(g_i^{(\ell)})$. The dot product and the norms then
concentrate over the $d$ coordinates, and because $g^{(\ell)}$ and $g^{(\ell')}$
are independent for $\ell\neq\ell'$,
\[
  m_\ell \cdot m_{\ell'} = \sum_i \sigma(g_i^{(\ell)})\,\sigma(g_i^{(\ell')})
  \;\to\; d\,\mathbb{E}[\sigma]^2,
  \qquad
  \|m_\ell\|^2 = \sum_i \sigma(g_i^{(\ell)})^2 \;\to\; d\,\mathbb{E}[\sigma^2],
\]
with $\mathbb{E}[\sigma]$, $\mathbb{E}[\sigma^2]$ the moments of $\sigma$ under a
standard Gaussian. Hence
\[
  \cos(m_\ell, m_{\ell'}) \;\to\;
  \frac{\mathbb{E}[\sigma]^2}{\mathbb{E}[\sigma^2]} \;=\; c_\ell .
\]
The cross term factors into $\mathbb{E}[\sigma]^2$ because the two layers'
pre-activations are independent: only the part common to every weight draw --- the
all-ones component, with per-coordinate value $\mathbb{E}[\sigma]$ --- survives the
product, and it carries a fraction $c_\ell$ of the energy $\mathbb{E}[\sigma^2]$.
This is the same $c_\ell$ that bounds the per-layer stable rank, $s_\sigma = 1/c_\ell$
(Table~\ref{tab:activations}): the per-layer spike and the cross-layer alignment
are one quantity. The collapse $z \approx \mathbf 1$ and the concentration of the
sums are leading-order statements, exact as the spike grows and $d \to \infty$.

\begin{figure*}[t]
  \centering
  \includegraphics[width=\textwidth]{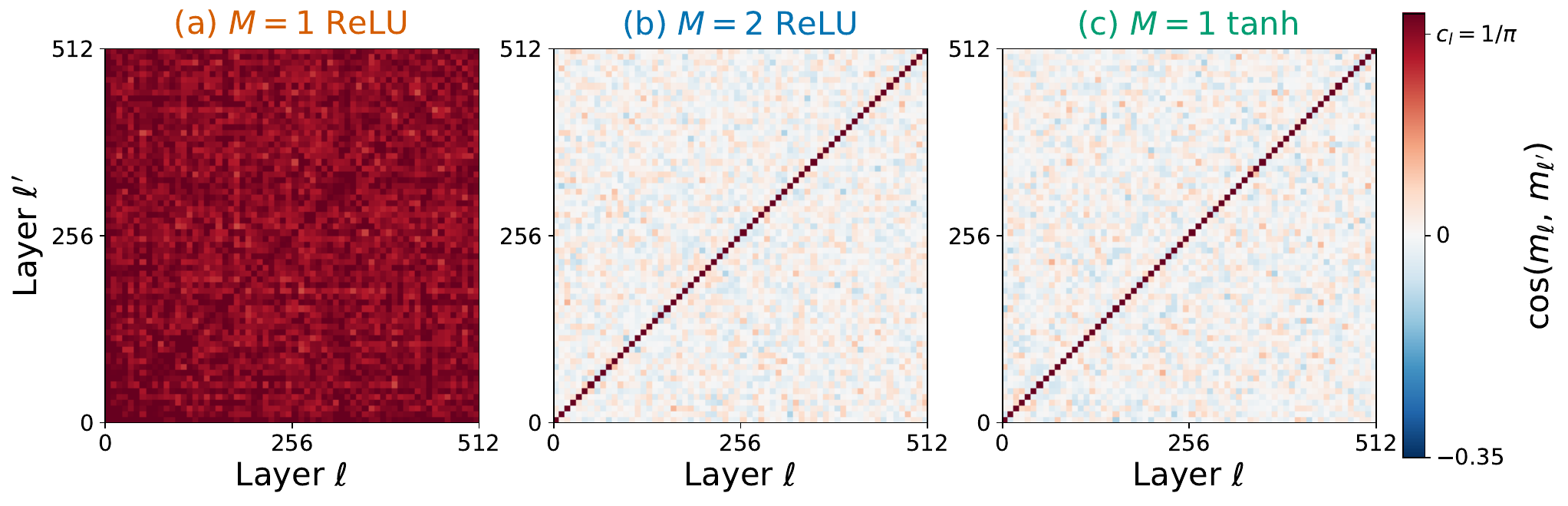}
  \caption{\textbf{The branch-mean correlation is the same for every pair of
  layers, not only neighbors.} Pairwise cosine $\cos(m_\ell, m_{\ell'})$ of the
  branch means at initialization (Pre-RMSNorm, $\beta{=}1$, $d{=}256$,
  $L{=}512$, $n{=}4096$, 3 seeds; the matrix samples the depth-$512$ network
  every $8$ layers).
  \textbf{(a)} A single uncentered matrix aligns every pair at the coherence
  constant $c_\ell = 1/\pi$ (off-diagonal mean $0.320$); the field is uniform, so
  the alignment does not decay with layer separation.
  \textbf{(b)} A second matrix decorrelates the means, leaving every
  off-diagonal pair at zero (mean $0.002$).
  \textbf{(c)} A centered activation has no mean to align (mean $-0.000$). The
  diagonal is trivially $1$.}
  \label{fig:coherence-heatmaps}
\end{figure*}

\paragraph{Anchor checks.}
The decomposition $m_\ell = \mathbb{E}[\sigma]\,\mathbf 1 + \varepsilon_\ell$, with $\varepsilon_\ell$
a zero-mean remainder independent across layers, makes two predictions we confirm
at initialization ($d{=}256$, $L{=}512$, $n{=}4096$, 3 seeds). Each branch mean
aligns with the all-ones direction at $\cos(m_\ell, \mathbf 1) \to \sqrt{c_\ell}$ as
depth grows, measured $0.565$ for ReLU and $0.433$ for GELU against
$\sqrt{1/\pi} = 0.564$ and $\sqrt{0.187} = 0.432$; the $\mathbf 1$-component
therefore carries a $\cos^2 = c_\ell$ fraction of the energy. The accumulated mean
then aligns fully, $\cos(\mu_\ell, \mathbf 1)\to 1$, as its $\mathbf 1$-part grows
$\propto\ell$ while the remainder grows only $\propto\sqrt\ell$. The cross-layer
cosine matches $c_\ell$ directly in Figure~\ref{fig:coherence-heatmaps}.

\paragraph{No backward analogue.}
The forward spike grows because each branch mean contains the weight-independent
component $\mathbb{E}[\sigma]\,\mathbf 1$, the same direction at every layer. The
backward injection has no such component. For a single matrix the branch
Jacobian is $J = \mathrm{diag}(\sigma'(W\hat z))\,W\,J_{\mathrm{norm}}$, so the
per-layer backward contribution is
\[
  n_\ell = \mathbb{E}_x\!\big[J^\top \delta_\ell\big]
         = J_{\mathrm{norm}}^\top\, W_\ell^\top\,
           \mathbb{E}_x\!\big[\sigma'(W_\ell\hat z)\odot \delta_\ell\big].
\]
Every term passes through $W_\ell^\top$, the layer's own weight matrix, so for
two layers $n_\ell$ and $n_{\ell'}$ carry $W_\ell^\top$ and $W_{\ell'}^\top$ from
independent weights: $\mathbb{E}[n_\ell^\top n_{\ell'}] = 0$ and
$\cos(n_\ell, n_{\ell'}) = O(1/\sqrt d)\to 0$. With no weight-independent shared
component, the injections of different layers are mutually random. (Two matrices
add a further rotation $W_2^\top$; a centered activation has no mean to rotate.)
The asymmetry is between the mean of $\sigma$, a fixed direction that is added,
and the gain $\sigma'$, a mask that only scales.

We confirm this at initialization ($d{=}256$, $L{=}512$, $n{=}512$, 3 seeds).
We start the backward recursion from two choices of $\delta_L$, the gradient of
the loss with respect to the final representation: a random Gaussian vector per
input, in place of any particular loss, and the exact gradient of a softmax
cross-entropy loss with a random ten-class head and random labels. In the three
branch configurations (one matrix, two matrices, and a centered activation),
the cosine similarity $\cos(n_\ell, n_{\ell'})$ between the backward
contributions of two layers averages $0.00$ over the layer pairs, and its
root-mean-square is $0.063$, which matches the random baseline
$1/\sqrt d = 0.0625$; the forward branch means $m_\ell$ of the one-matrix ReLU
branch align at $c_\ell = 0.32$. Under the random $\delta_L$, the $n \times d$
matrix of gradients $\delta_\ell$ preserves its rank
($\mathrm{erank}/d \ge 0.72$, $\approx 0.9$ for one-matrix ReLU) and grows no
mean spike (mean fraction $\approx 0.002$; the forward mean fraction reaches
$1.0$). Under the cross-entropy loss, $\delta_L$ lies in the row space of the
ten-class head, so the gradient matrix is effectively rank ten before any layer
acts on it. This independence of the per-layer weights is a property of
initialization; we do not analyze the trained network.

\paragraph{Pairwise alignment across all layer pairs
(Figure~\ref{fig:coherence-heatmaps}).}
The headline result reports the correlation between the branch means of any two
layers, while the consecutive-layer panel of Figure~\ref{fig:coherence}a shows
only neighbors. Figure~\ref{fig:coherence-heatmaps} plots the full matrix
$\cos(m_\ell, m_{\ell'})$, which is flat: for a single uncentered matrix every
pair correlates at $c_\ell$ (off-diagonal mean $0.320$ for ReLU against
$1/\pi = 0.318$), and for two matrices or a centered activation every pair sits
at zero (off-diagonal mean $0.002$ and $-0.000$ respectively). The correlation
does not fall off with the layer separation $|\ell - \ell'|$ because it is
carried by one shared direction common to every layer rather than by a chain of
layer-to-layer dependences: the all-ones component sits in each branch mean
regardless of that layer's weights, so the alignment is the same at every
separation. The main text therefore treats the cross-layer correlation as a
single constant $c_\ell$.

\paragraph{Mean-growth exponents and the probe-size control (Figure~\ref{fig:coherence}b).}
The dashed guides are power laws $\|\mu_\ell\| \propto \ell^{q}$ anchored at the
$\ell{=}1$ value of each curve, with $q{=}1$ for the coherent single uncentered
branch and $q{=}\tfrac{1}{2}$ otherwise. A finite probe of $n$ inputs estimates a
zero mean only up to sampling noise: for a zero-mean process the expected
measured value is $\sqrt{\mathbb{E}\|z_\ell\|^2/n}$ (the gray noise floor in the
figure), which random-walks across layers at any $n$ and shrinks only as
$1/\sqrt n$; we therefore vary $n$ to separate real means from noise. Fitted
over $\ell\in[50,512]$, M=1 ReLU ($q{=}0.99$, $\propto\ell$) and M=2 ReLU
($q{=}0.56$, $\propto\sqrt\ell$) are $n$-independent, hence real. SwiGLU likewise has a small but $n$-independent mean
($q{\approx}0.5$): a gate \emph{decorrelates} the mean into a random walk rather
than removing it. tanh's apparent $\sqrt\ell$ instead scales as $1/\sqrt n$ and
vanishes as $n{\to}\infty$, confirming that a truly centered activation has zero
mean; in the figure its curve coincides with the floor. Probe sizes in
Figure~\ref{fig:coherence}: the mean-magnitude panel uses $n{=}32768$ so the small
real means sit well above the floor; the alignment panel uses $n{=}4096$; the
rank and spectrum panels are $n$-independent for $n{>}d$ and use $n{=}512$.

\paragraph{Width independence (Table~\ref{tab:width}).}
We confirm the collapse is not specific to narrow networks.
Table~\ref{tab:width} reports the effective rank $\mathrm{erank}(Z_\ell)/d$ for
M=1 ReLU at four widths from $128$ to $1024$: the trajectory is nearly
width-independent, falling to a few percent of the width by depth $128$ at every
width, and the mean fraction reaches $1.00$ by the same depth.

\begin{table}[t]
  \centering
  \caption{\textbf{The rank collapse is not a narrow-network artifact.}
  Effective rank as a fraction of width, $\mathrm{erank}(Z_\ell)/d$, for M=1
  ReLU at initialization (Pre-RMSNorm, $\beta{=}1$, mean over 3 seeds,
  $n=\max(1024,2d)$ probe inputs). The collapse occurs at every width.}
  \label{tab:width}
  \begin{tabular}{lrrrrr}
    \toprule
    & \multicolumn{5}{c}{Layer $\ell$} \\
    \cmidrule(lr){2-6}
    Width $d$ & $32$ & $64$ & $128$ & $256$ & $512$ \\
    \midrule
    $128$  & $0.216$ & $0.097$ & $0.044$ & $0.023$ & $0.015$ \\
    $256$  & $0.229$ & $0.100$ & $0.040$ & $0.018$ & $0.010$ \\
    $512$  & $0.250$ & $0.109$ & $0.040$ & $0.015$ & $0.007$ \\
    $1024$ & $0.296$ & $0.136$ & $0.047$ & $0.015$ & $0.006$ \\
    \bottomrule
  \end{tabular}
\end{table}

\begin{figure*}[t]
  \centering
  \includegraphics[width=\textwidth]{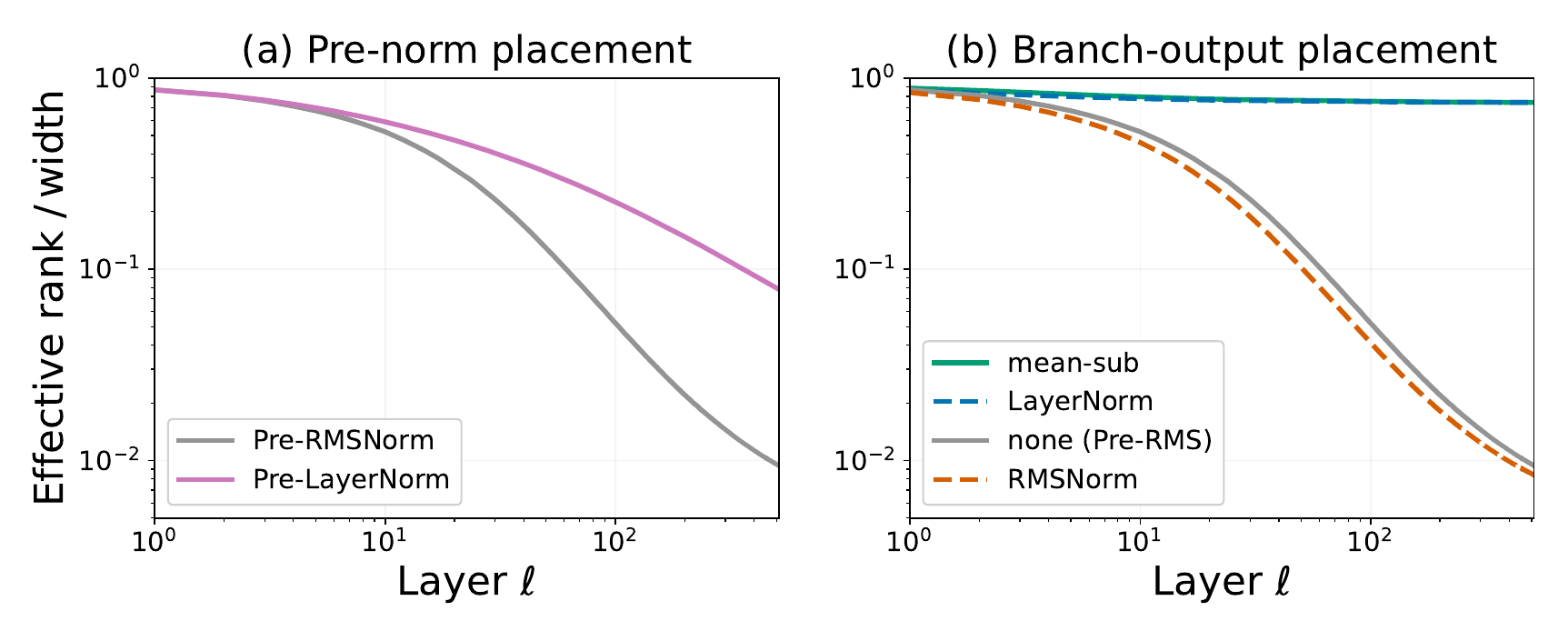}
  \caption{\textbf{Pre-Norm cannot remove the mean spike; a branch-output
  normalization can, but only by subtracting the mean.} Effective rank
  $\mathrm{erank}(Z_\ell)/d$ for M=1 ReLU at initialization ($\beta{=}1$,
  $d{=}256$, 3 seeds).
  \textbf{(a)} In the Pre-Norm placement both normalizers collapse:
  Pre-LayerNorm, which centers the branch input, reaches $0.078$ by depth $512$
  against $0.009$ for Pre-RMSNorm --- roughly eightfold slower, but still
  collapsed.
  \textbf{(b)} At the branch output (Pre-RMSNorm fixed), mean subtraction and
  LayerNorm preserve the rank ($0.745$), while RMSNorm and no branch
  normalization collapse ($0.008$ and $0.009$): the centering matters, the
  rescaling does not.}
  \label{fig:coherence-norm}
\end{figure*}

\paragraph{Normalization placement (Figure~\ref{fig:coherence-norm}).}
To understand how normalization placement interacts with this mean spike, we
measure the effective rank of the representation under a Pre-Norm and under a
branch-output normalization, and find that only a branch-output normalization
that subtracts the mean prevents the collapse (Figure~\ref{fig:coherence-norm}).
In the Pre-Norm placement the rank collapses regardless of the normalizer
(panel a): centering the branch input with LayerNorm slows the collapse relative
to RMSNorm, because it breaks the feedback loop by which the stream's accumulated
mean re-enters the branch, but the rank still collapses because the activation
re-creates the mean at its output. At the branch output, after the activation,
the rank is preserved only if the normalization subtracts the mean (panel b):
mean subtraction and LayerNorm hold it high, while RMSNorm, which only rescales,
collapses identically to no normalization.

Peri-LN \citep{kim2025peri} proposes a normalization at the branch-output
placement in addition to the Pre-Norm. Their experiments use two matrices per
branch, so they do not encounter this single-matrix spike.

\begin{figure*}[t]
  \centering
  \includegraphics[width=\textwidth]{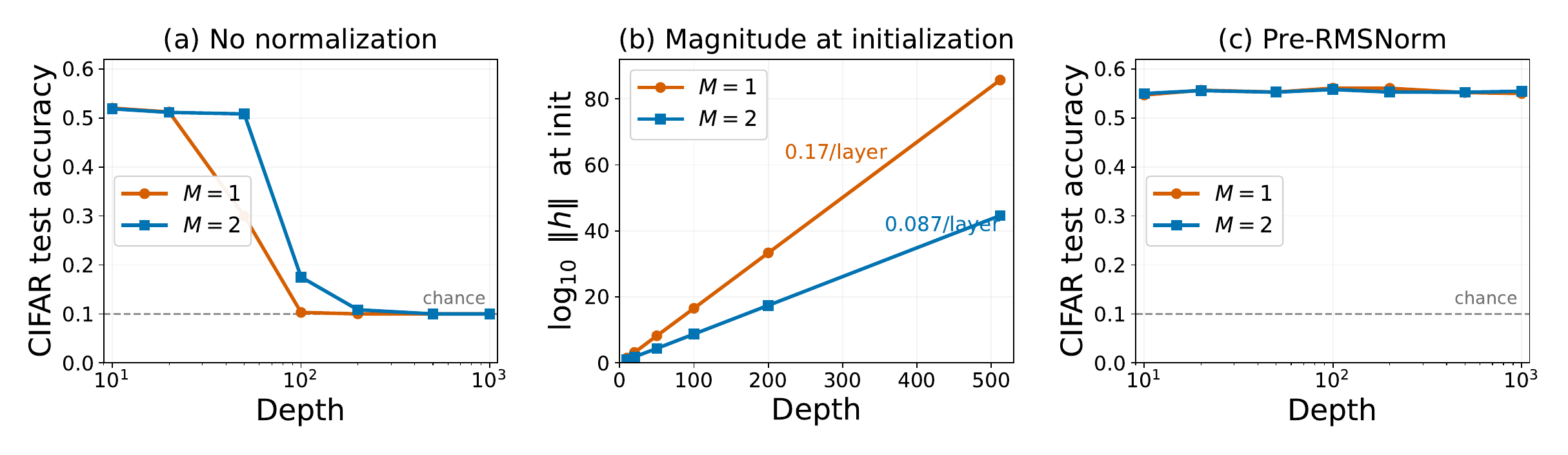}
  \caption{\textbf{Without normalization the coherent mean spike halves the
  trainable depth; with normalization the rank collapse it causes does not
  impair CIFAR training.} All ReLU, $\beta{=}1$, width $256$.
  \textbf{(a)} Without normalization, the single-matrix architecture ($M{=}1$)
  fails by about depth $50$ and the two-matrix architecture ($M{=}2$) by about
  depth $100$ (best test accuracy over the learning-rate sweep; CIFAR chance
  accuracy $0.1$).
  \textbf{(b)} The residual magnitude at initialization grows exponentially with
  depth, twice as fast for the single-matrix architecture ($0.17$ versus $0.087$
  decades per layer), so it reaches a given magnitude at half the depth.
  \textbf{(c)} With Pre-RMSNorm the magnitudes are controlled and the two
  architectures reach similar best accuracy over the learning rate sweep
  (within a point at every depth up to $1000$), even though the single-matrix
  representation has collapsed to rank $\approx 2$ at initialization
  (Figure~\ref{fig:coherence}d).}
  \label{fig:coherence-training}
\end{figure*}

\paragraph{Training scope: the magnitude and rank channels
(Figure~\ref{fig:coherence-training}).}
We train single-matrix ($M{=}1$) and two-matrix ($M{=}2$) ReLU networks on
CIFAR-10 with Adam for $5000$ steps (width $256$, batch $256$, $\beta{=}1$),
sweeping the learning rate over six values from $3{\times}10^{-5}$ to $10^{-2}$
(the two largest omitted at depths $500$ and $1000$, where the optimal rate is
smaller) and reporting the best test accuracy over the sweep. The one
systematic difference is at the smallest learning rate tested
($3{\times}10^{-5}$, well below the optimum of $10^{-3}$), where the
single-matrix architecture trains 4--5 points higher at depths ${\geq}100$;
this is consistent with an artifact of the fixed step budget, since the
single-matrix network has half as many weight matrices to train at a learning
rate this small. The
initialization magnitude in panel~(b) is the accumulated $\log_{10}$ of the
residual RMS; the per-layer renormalization that keeps it finite at large depth
is exact for ReLU, so the two slopes are measured rather than fit.

\section{Width Expansion: Full Marchenko--Pastur Derivation}
\label{app:mp}

This appendix derives the results used in Section~\ref{sec:width}.
Sections~\ref{app:mp-jacobian}--\ref{app:mp-threshold} treat the exact case: we
factorize the branch Jacobian, reduce the surviving submatrix to a
Marchenko--Pastur ensemble for ReLU and absolute value, and read off the rank
threshold $m/d = 1/p(\sigma)$ and its conditioning. The later subsections cover
smooth activations, the freeness assumption and its validation in trained
models, the measured spectra of all four activations, the branch Jacobian's own
conditioning, and the Gaussian-product reference curve of
Figure~\ref{fig:width-cond}.

\subsection{Jacobian factorization and the survival rate}
\label{app:mp-jacobian}

The feedforward branch $f(x) = W_{\mathrm{down}}\,\sigma(W_{\mathrm{up}} x)$, with
$W_{\mathrm{up}} \in \R^{m\times d}$ and $W_{\mathrm{down}} \in \R^{d\times m}$,
has Jacobian
\begin{equation}
  J_f(x) = W_{\mathrm{down}}\, D(x)\, W_{\mathrm{up}}, \qquad
  D(x) = \diag\!\big(\sigma'(W_{\mathrm{up}} x)\big).
  \label{eq:mp-jac}
\end{equation}
The mask $D(x)$ is diagonal, so it can only lower rank:
$\rank J_f \le \#\{i : \sigma'((W_{\mathrm{up}} x)_i) \neq 0\}$. We call the
expected fraction of units with nonzero derivative the survival rate,
$p(\sigma) = \Pr_{z\sim\mathcal{N}(0,1)}[\sigma'(z)\neq 0]$. ReLU has
$\sigma'=\mathbf{1}[z>0]$ and $p=\tfrac12$; absolute value has
$\sigma'=\mathrm{sign}(z)$ and $p=1$. At initialization the pre-activations are
i.i.d.\ and mean-zero, so $D$ keeps $pm$ rows of $W_{\mathrm{up}}$ in expectation.

\subsection{The surviving submatrix}
\label{app:mp-submatrix}

For ReLU and absolute value the derivative takes only the values $0$ and $\pm1$,
so $D$ either drops a row of $W_{\mathrm{up}}$ or keeps it at unit magnitude; it
never rescales the surviving rows, which is what makes these two activations
exact.

For absolute value, $D = \diag(\mathrm{sign}(W_{\mathrm{up}} x))$ flips the signs
of some rows, and the signs cancel in the Gram,
$(D\,W_{\mathrm{up}})^{\top}(D\,W_{\mathrm{up}}) = W_{\mathrm{up}}^{\top}W_{\mathrm{up}}$.
The surviving submatrix has exactly the spectrum of the full Gaussian
$W_{\mathrm{up}}$, with aspect ratio $\gamma = m/d$.

For ReLU the surviving rows $\{w_i : w_i^{\top} x > 0\}$ are Gaussian rows
restricted to a half-space. The conditioning gives each a mean
$\sqrt{2/\pi}\,\hat{x}$ along $\hat{x}=x/\|x\|$ and shrinks its variance there
from $1$ to $1-\tfrac{2}{\pi}$, while the other $d-1$ directions stay standard
Gaussian. What Marchenko--Pastur needs is only the matrix the Gram converges to,
the second moment $\E[w_i w_i^{\top}\mid w_i^{\top}x>0]$, and there the reduced
variance and the squared mean cancel along $\hat{x}$:
\begin{equation}
  \E\!\big[w_i w_i^{\top}\mid w_i^{\top}x>0\big]
  = \underbrace{\big(I - \tfrac{2}{\pi}\hat{x}\hat{x}^{\top}\big)}_{\text{covariance}}
  + \underbrace{\tfrac{2}{\pi}\hat{x}\hat{x}^{\top}}_{(\text{mean})(\text{mean})^{\top}}
  = I.
  \label{eq:mp-second-moment}
\end{equation}
The Gram therefore converges to the identity, and the surviving submatrix is an
i.i.d.\ ensemble with second moment $I$ and aspect ratio
$\gamma = pm/d = \tfrac12\,(m/d)$.

\subsection{Marchenko--Pastur law and the full-rank threshold}
\label{app:mp-threshold}

In the proportional limit $d, pm \to \infty$ with $\gamma = pm/d$ fixed, the
eigenvalues of $\tfrac{1}{pm}(D\,W_{\mathrm{up}})^{\top}(D\,W_{\mathrm{up}})$
converge to the Marchenko--Pastur law \citep{marchenko1967distribution} on
$[\lambda_-,\lambda_+]$ with
$\lambda_{\pm} = (1 \pm 1/\sqrt{\gamma})^2$. The lower edge $\lambda_-$ is zero at
$\gamma=1$ and positive for $\gamma>1$, so the submatrix is full rank if and only
if $\gamma \ge 1$, i.e.\ $m/d \ge 1/p(\sigma)$. This threshold $m/d = 1/p$ equals
$2$ for ReLU and $1$ for absolute value. At the threshold the submatrix is full
rank but its smallest eigenvalue reaches zero, so it is badly conditioned; for
$\gamma>1$ the spectrum separates from zero and the condition number
$\sigma_{\max}/\sigma_{\min} = (1+1/\sqrt\gamma)/(1-1/\sqrt\gamma)$ is finite,
falling from $\infty$ at $\gamma=1$ to $5.8$ at $\gamma=2$ and $3.0$ at $\gamma=4$
(the ReLU row of Figure~\ref{fig:width-spectrum}, at $m/d = 2,4,8$).

The branch Jacobian inherits this rank. Because
$W_{\mathrm{down}}\in\R^{d\times m}$ with $m \ge d$ is full rank and independent
of $D\,W_{\mathrm{up}}$ at initialization, the column space of $D\,W_{\mathrm{up}}$
(of dimension $\rank(D\,W_{\mathrm{up}}) \le d$) meets the $(m-d)$-dimensional
null space of $W_{\mathrm{down}}$ only at the origin, so
$\rank J_f = \rank(D\,W_{\mathrm{up}})$. The width threshold for the branch
Jacobian is the same as for the surviving submatrix.

\subsection{Smooth activations}
\label{app:mp-smooth}

For GELU and SiLU the derivative is never exactly zero, so
$D=\diag(\sigma'(W_{\mathrm{up}}x))$ rescales every row rather than dropping any.
The Gram becomes
$\tfrac{1}{n_{\mathrm{eff}}}W_{\mathrm{up}}^{\top}\Lambda\,W_{\mathrm{up}}$ with
$\Lambda=\diag(\sigma'(W_{\mathrm{up}}x)^2)$ and
$n_{\mathrm{eff}}=\sum_i\sigma'(z_i)^2$: a Gaussian matrix with a per-row variance
profile, not the masked submatrix in Section~\ref{app:mp-submatrix}.

This $n_{\mathrm{eff}}$ is the effective number of surviving rows, equal to the
count $pm$ when $\sigma'^2\in\{0,1\}$. The aspect ratio is then
$\gamma = n_{\mathrm{eff}}/d \to \E[\sigma'^2]\,(m/d)$, so the effective survival
rate is the mean squared derivative $\E[\sigma'^2]$: $p$ for ReLU and absolute
value, $0.46$ and $0.38$ for GELU and SiLU.

The spectrum is no longer exactly Marchenko--Pastur: a variance profile deforms
the law by the whole distribution of $\sigma'^2$, not its mean
alone \citep{pennington2017nonlinear}. Standard Marchenko--Pastur at
$\gamma=\E[\sigma'^2]\,(m/d)$ matches the scale but not the deformation, a close
but inexact fit, and no single scalar survival rate reproduces the spectrum
exactly. We use $\E[\sigma'^2]$, the mean of the weight distribution and the one
available in closed form.

\subsection{Free-bulk assumption and trained weights}
\label{app:mp-freeness}

The effective-rank prediction in Figure~\ref{fig:width-cond} composes the
spectrum of $D\,W_{\mathrm{up}}$ with $W_{\mathrm{down}}$, which requires the two
matrices to be asymptotically free: their singular bases in the shared
$m$-dimensional hidden space should show no systematic alignment. This holds
exactly at initialization, where the two are independent Gaussians.

In pretrained GPT-2 Medium we measure the coherence between these bases,
normalized so that free bases give $1$ (Figure~\ref{fig:width-cond}c). Averaged
over all directions it stays close to free across the 24 FFN blocks, with a mean
of $0.96$; the first two blocks are higher ($1.6$ and $1.3$) and the rest fall
between $0.78$ and $1.02$. The most- and least-aligned twenty directions deviate
more (a low-rank trained signal, with coherence near $5$ at the top and $3$ at
the bottom), but they are a small fraction of the $1024$ directions and leave the
bulk that determines the effective rank approximately free.
\citet{daneshmand2021bn} prove an analogous orthogonalization for BatchNorm;
whether it extends to LayerNorm is open, and these measurements are direct
evidence that the bulk is close to free in a trained transformer.

\subsection{Spectra for all four activations}
\label{app:mp-grid}

\begin{figure*}[t]
  \centering
  \includegraphics[width=0.85\textwidth]{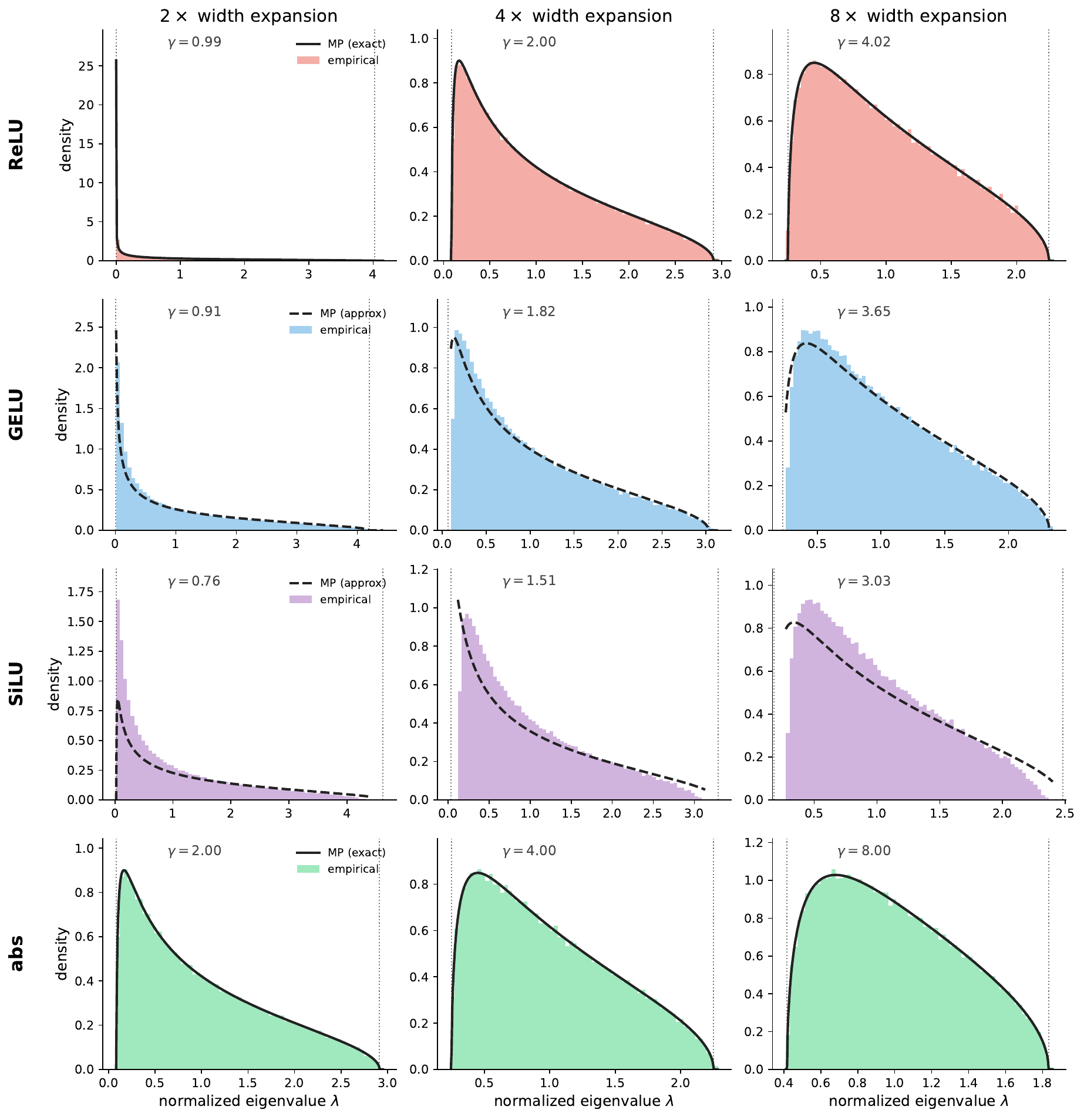}
  \caption{\textbf{Submatrix spectra for all four activations.} Measured Gram
  eigenvalue histograms (filled) of $D\,W_{\mathrm{up}}$ against the
  Marchenko--Pastur density (line), for ReLU, GELU, SiLU, and abs (rows) at width
  expansions $m/d \in \{2,4,8\}$ (columns). Exact (solid) for ReLU and abs,
  approximate (dashed) for GELU and SiLU. Companion to
  Figure~\ref{fig:width-spectrum}, which omits SiLU. Random Gaussian
  $W_{\mathrm{up}}$ at initialization, $d = 512$, $40$ seeds.}
  \label{fig:width-spectrum-app}
\end{figure*}

Figure~\ref{fig:width-spectrum-app} overlays the measured Gram eigenvalues on
the Marchenko--Pastur density for all four activations at $m/d\in\{2,4,8\}$. The
fit is exact for ReLU and absolute value (solid) and approximate for GELU and
SiLU (dashed), as Sections~\ref{app:mp-submatrix}--\ref{app:mp-smooth} predict.
SiLU, deferred from Figure~\ref{fig:width-spectrum}, follows the GELU pattern.
The smooth-activation deformation in Section~\ref{app:mp-smooth} is visible at
small $m/d$, largest for SiLU at $m/d = 2$, and shrinks as $m/d$ grows. At
$m/d = 2$ the smooth activations sit below the full-rank threshold
($\gamma = p\,(m/d) < 1$), so the Gram spectrum carries a point mass at zero of
weight ${\approx}\,1 - \gamma$, visible as the spike in the first histogram
bins, which the continuous Marchenko--Pastur overlay does not include.

\subsection{Conditioning of the branch Jacobian itself}
\label{app:mp-jf-cond}

Above the threshold the submatrix is well-conditioned, but the branch Jacobian
$J_f = W_{\mathrm{down}}(D\,W_{\mathrm{up}})$ is not. Restricted to the
$d$-dimensional column space of $D\,W_{\mathrm{up}}$, $W_{\mathrm{down}}$ acts as
a square $d\times d$ Gaussian, which is poorly conditioned; widening the branch
enlarges $m$ but leaves this $d\times d$ map unchanged.

The submatrix conditioning improves with width (Figure~\ref{fig:width-spectrum})
but the conditioning of the Jacobian $J_f$ does not: for ReLU at $d=512$ (weights
at the default $1/\text{fan-in}$ variance), as
$\gamma$ grows from $2$ to $16$ the submatrix's smallest Gram eigenvalue rises
from $0.09$ to $0.57$, while $J_f$'s stays near $10^{-6}$ even where $J_f$ is full
rank. The effective rank of $J_f$, however, does improve with width, and the
residual skip connection keeps the propagated gradient full rank.

\subsection{The Gaussian-product reference curve}
\label{app:mp-product}

The reference curve in Figure~\ref{fig:width-cond}b is the effective rank per
dimension, $\mathrm{erank}(J_f)/d$, of a product of two Gaussian matrices
($W_{\mathrm{up}}$ and $W_{\mathrm{down}}$ both Gaussian, with no activation) as a
function of $\gamma$. We compute it numerically: at each $\gamma$ we draw the two
factors, form the product, and average $\mathrm{erank}(J_f)/d$ over seeds
($d=256$). The curve rises with $\gamma$ and reaches $\approx 0.80$ over the
plotted range.

It sits below the surviving submatrix's effective rank at every $\gamma$ because
the product spreads the spectrum (the same structure behind $J_f$'s poor
conditioning in Section~\ref{app:mp-jf-cond}), so the branch Jacobian never reaches the
effective rank of a single Gaussian matrix, even where it is full rank.

\end{document}